\documentclass[10pt,twocolumn,letterpaper]{article}

\usepackage{cvpr}              
\definecolor{cvprblue}{rgb}{0.21,0.49,0.74}
\usepackage[pagebackref,breaklinks,colorlinks,allcolors=cvprblue]{hyperref}

\usepackage{algorithm}
\usepackage{algorithmic}
\usepackage{multirow}
\usepackage{amssymb}
\usepackage{siunitx}

\def\paperID{} 
\def\confName{CVPR}
\def\confYear{2026}

\title{B$^3$-Seg: Camera-Free, Training-Free 3DGS Segmentation via Analytic EIG and Beta–Bernoulli Bayesian Updates}

\author{Hiromichi Kamata\\
Sony Group Corporation\\
Tokyo, Japan\\
{\tt\small Hiromichi.Kamata@sony.com}
\and
Samuel Arthur Munro\\
Pixomondo\\
London, United Kingdom\\
{\tt\small samuelm@pixomondo.com}
\and
Fuminori Homma\\
Sony Group Corporation\\
Tokyo, Japan\\
{\tt\small Fuminori.Homma@sony.com}
}

\begin{document}
\maketitle
\begin{abstract}
Interactive 3D Gaussian Splatting (3DGS) segmentation is essential for real-time editing of pre-reconstructed assets in film and game production.
However, existing methods rely on predefined camera viewpoints, ground-truth labels, or costly retraining, making them impractical for low-latency use.
We propose \textbf{B$^3$-Seg (Beta--Bernoulli Bayesian Segmentation for 3DGS)}, a fast and theoretically grounded method for open-vocabulary 3DGS segmentation under \textbf{camera-free} and \textbf{training-free} conditions.
Our approach reformulates segmentation as sequential Beta--Bernoulli Bayesian updates and actively selects the next view via analytic Expected Information Gain (EIG).
This Bayesian formulation guarantees the adaptive monotonicity and submodularity of EIG, which produces a greedy $(1{-}1/e)$ approximation to the optimal view sampling policy.
Experiments on multiple datasets show that B$^3$-Seg achieves competitive results to high-cost supervised methods while operating end-to-end segmentation within a few seconds.
The results demonstrate that B$^3$-Seg enables practical, interactive 3DGS segmentation with provable information efficiency.
\end{abstract}   

\section{Introduction}
Recently, 3D Gaussian Splatting (3DGS) \cite{3dgs} has rapidly gained attention as a 3D representation that combines real-time rendering with high visual fidelity. 
In film and game production, it is increasingly common that only a pre-reconstructed 3DGS asset is shared across teams. 
Thus, interactive 3DGS segmentation—selecting, editing, and removal of objects directly on the asset—has become a required capability \cite{nerfstudio, supersplat}.

Many recent works tackle open-vocabulary 3DGS segmentation and achieve strong accuracy \cite{langsplat, gaussian-grouping, opengaussian, objectgs, ludvig}. 
However, most methods assume access to predefined camera viewpoints and/or ground-truth semantic masks, which does not align with the practical setting where only a shared 3DGS asset is available. 
In real use, interactive editing must operate camera-free, training-free, and open-vocabulary, and return results within a few seconds.
\begin{figure}[t]
    \centering
    \includegraphics[width=0.95\columnwidth]{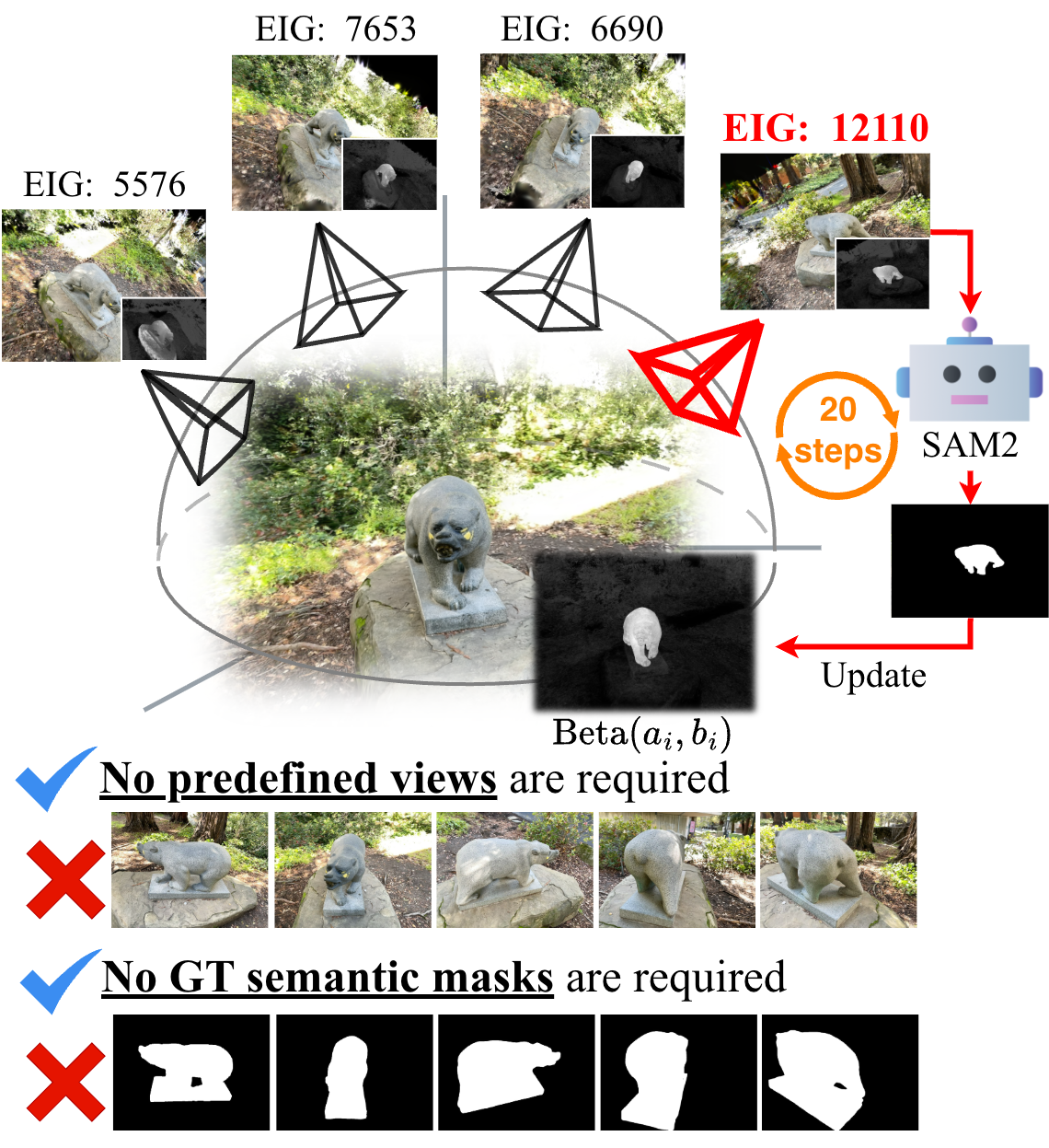} 
\caption{B$^3$-Seg actively selects the next view via Expected Information Gain and updates 3D labels by Beta--Bernoulli Bayesian updates. It runs in a few seconds and requires neither predefined views nor ground-truth semantic labels.}
\label{fig:autoreg}
\end{figure}

Interactive editing also requires low latency. 
Although several recent methods \cite{gaussian-grouping, opengaussian, ludvig, objectgs} are highly accurate, they are based on large-scale pretraining and often take minutes to tens of minutes per scene. 
Faster approaches that return results within a minute have also appeared \cite{flashsplat, cob-gs}, but still assume access to camera viewpoints and ground-truth labels.

We introduce \textbf{B$^3$-Seg (Beta–Bernoulli Bayesian Segmentation for 3DGS)}. 
B$^3$-Seg runs under \textbf{camera-free, training-free, open-vocabulary} conditions and produces results within \textbf{a few seconds}. 
We reformulate 3DGS segmentation as sequential Beta–Bernoulli Bayesian updates and select the next best view via analytic Expected Information Gain (EIG). 
On each actively selected view, we obtain masks using Grounding~DINO + SAM2 \cite{grounded-sam} with CLIP \cite{clip} re-ranking, and we update the Beta posteriors accordingly. 
Building on this Bayesian formulation, we prove the non-negativity and diminishing returns of EIG, and derive a greedy $(1{-}1/e)$ approximation guarantee. 
This yields an interactive 3DGS segmentation method that is both information-efficient and theoretically grounded.

Our contributions are as follows.
\begin{itemize}
\item \textbf{Camera-free, training-free, and few-second segmentation.} We achieve open-vocabulary 3DGS segmentation in a few seconds without camera viewpoints, ground-truth labels, or retraining.
\item \textbf{Bayesian reformulation.} We reformulate 3DGS segmentation as sequential Beta--Bernoulli Bayesian updates, providing a unified and robust probabilistic model.
\item \textbf{Analytic EIG and active view selection.} We estimate per-view pseudo-counts from one render and choose the next view using analytic EIG.
\item \textbf{Theoretical guarantees.} We show non-negativity and diminishing returns of EIG, confirming adaptive monotonicity and adaptive submodularity, leading to greedy $(1{-}1/e)$ approximation.
\item \textbf{Competitive accuracy.} Across datasets, our method is competitive with slower, label-dependent baselines while preserving the practical constraints above.
\end{itemize}

\section{Related Works}

\subsection{3DGS Segmentation}
Recent works improve 3DGS with semantics through text conditioning and cross-modal supervision, allowing open-vocabulary retrieval and segmentation \cite{langsplat, opengaussian, lerf, gaussian-grouping, objectgs}.
LERF \cite{lerf} integrates CLIP features into a NeRF by optimizing a dense language field across views, enabling flexible text queries but requiring access to multi-view images and extended scene optimization.
Gaussian Grouping \cite{gaussian-grouping} effectively groups Gaussians into instances/parts using multi-view supervision and feature aggregation, achieving high accuracy with additional optimization for all reconstruction views. 
OpenGaussian \cite{opengaussian} distills open-vocabulary 2D segmenters into 3D Gaussians using cross-modal pretraining and multi-stage pipelines with multi-view images and camera trajectories.
ObjectGS \cite{objectgs} creates an object-centric 3DGS with instance-level reasoning and compositional editing. However, it requires strong semantic priors, and extensive computation for per-scene optimization.
These methods achieve high accuracy, but their need for reconstruction images/camera paths, ground-truth or distilled labels, and significant optimization time limits their use in interactive editing with only a stand-alone 3DGS asset.

\subsection{Few-Second 3DGS Segmentation}
To improve responsiveness, few-second frameworks such as FlashSplat \cite{flashsplat} and COB-GS have been proposed \cite{cob-gs}. 
FlashSplat casts the consistency between 2D masks and 3D labels as a linear program, and COB-GS maintains quality via boundary-aware improvements and texture heuristics. 
However, these approaches still assume the availability of reconstruction views and semantic masks, and they provide no theoretical guarantees. iSegMan~\cite{isegman} presents an interactive segmentation framework for 3DGS scenes. It uses visibility voting from multiple views based on empirically designed camera trajectories.

\subsection{Active View Sampling on 3DGS}
Active view selection has been widely studied for 3D reconstruction. 
FisherRF \cite{fisherrf} formulates the next-best view using Fisher information. 
ActiveGS \cite{activegs} uses 3DGS as a map representation to greedily select the next-best view based on confidence and coverage. 
ActiveGAMER \cite{activegamer} leverages fast 3DGS rendering and an information-gain objective with coarse-to-fine search to improve fidelity and completeness. 
ActiveSGM \cite{activesgm} integrates active camera sampling and segmentation in indoor scenes such as Replica \cite{replica}, but requires backpropagation-based training and targets a fixed set of classes via OneFormer \cite{oneformer}. 
In contrast, our method provides Bayesian updates with analytic EIG, and operates camera-free and training-free under open-vocabulary conditions.

\begin{figure*}[t]
    \centering
    \includegraphics[width=0.95\linewidth]{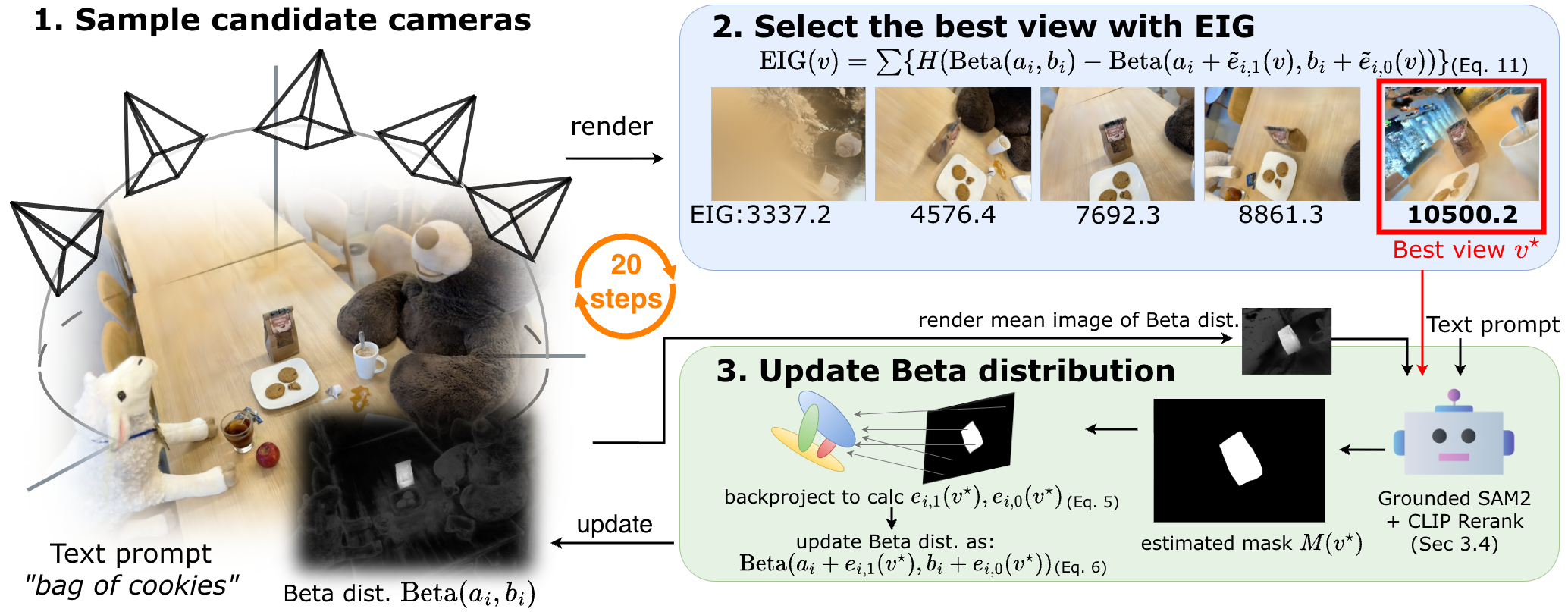} 
    \caption{\textbf{Overview of B$^3$-Seg}. 
(1) Sample $N_{\mathrm{cand}}$ candidate views on a sphere centered at the estimated object center $\mathbf{c}_{\mathrm{obj}}$.
(2) Render each candidate to compute $\mathrm{EIG}(v)$ by Eq.~\eqref{eq:eig}, and pick the best view $v^\star$ (red).
On $v^\star$, obtain masks using Grounded~SAM2 and CLIP reranking.
(3) From the mask, compute $(e_{i,1},e_{i,0})$ by Eq.~\eqref{eq:agg-evidence} and update Beta parameters.
We iterate (1)–(3) process 20 steps.
The pipeline enables camera-free, training-free, open-vocabulary 3DGS segmentation in a few seconds.}
    \label{fig:method}
\end{figure*}

\section{Proposed Method}

B$^3$-Seg performs a few-second and accurate segmentation of a user-specified object from a 3DGS scene under camera-free and training-free conditions. 
The method has two components: 
(i) 3DGS segmentation via sequential Beta–Bernoulli Bayesian updates of per-Gaussian probabilities, and (ii) active view selection based on the EIG in Beta distributions. 

\subsection{Preliminaries}
\label{sec:preliminaries}
3DGS \cite{3dgs} represents a scene as a set of Gaussians $\mathcal{G}{=}\{g_i\}_{i=1}^N$. Each Gaussian $g_i$ has mean $\mu_i$, covariance $\Sigma_i$, opacity $\alpha_i$, and color $c_i$. Rendering from a camera view $v$ accumulates per-Gaussian contributions as
\begin{align}
    I(v) = \sum_i c_i \, \alpha_i \, T_i,
\end{align}
where $T_i = \prod_{j=1}^{i-1}(1-\alpha_j)$ denotes the transmittance along the ray in view $v$. 

FlashSplat \cite{flashsplat} uses linear programming to solve binary assignments $P_i\in\{0,1\}$, determining whether $g_i$ is part of the target object or not.
\begin{equation}
\min_{\{P_i\}}\;
\sum_{v}\;\sum_{(j,k)\in I(v)}
\Bigl|\;\sum_{i} P_i\,\alpha_i\,T_i\;-\;M_{j,k}(v)\;\Bigr|,
\label{eq:fs-lp}
\end{equation}
where $M(v)$ is the binary mask of $I(v)$. 
FlashSplat produces the following decision rule:
\begin{align}    
&P_i = {\arg\max}_n A_{i,n}, \quad n \in \{0,1\}, \nonumber\\
\mathrm{where}&\quad A_{i,n} = \sum_v\sum_{(j,k)\in I(v)} \alpha_i T_i\,\mathbb{I}[M_{j,k}(v)=n],
\label{eq:fs-solve}
\end{align}
which compares visible responsibility inside and outside the mask.
COB-GS \cite{cob-gs} uses a similar gradient descent update. 

\subsection{Bayesian Reformulation of 3DGS Segmentation}
\label{sec:bayesian-reformulation}
We use the insights of Sec \ref{sec:preliminaries} to reformulate 3DGS segmentation as probabilistic Bayesian updates.
Let $y_i\in\{0,1\}$ indicate whether Gaussian $g_i$ belongs to the user-specified object. 
We place a Bernoulli–Beta prior/posterior on $p_i=\Pr(y_i{=}1)$:
\begin{align}
y_i\mid p_i\sim \mathrm{Bernoulli}(p_i), \qquad p_i\sim\mathrm{Beta}(a_i,b_i).
\end{align}
Given a rendered image $I(v)$ and an object mask $M(v)$, we treat per-pixel responsibilities as observations of success counts $e_{i,1}(v)$ and failure counts $e_{i,0}(v)$:
\begin{align}
e_{i,1}(v) &= \sum_{(j,k) \in I(v)} \alpha_i T_i \, \mathbb I[M_{j,k}(v)=1], \nonumber\\
e_{i,0}(v) &= \sum_{(j,k) \in I(v)} \alpha_i T_i \, \mathbb I[M_{j,k}(v)=0]. 
\label{eq:agg-evidence}
\end{align}
By Beta-Bernoulli conjugacy, the posterior updates are
\begin{align}
 \mathrm{Beta}(a_i, b_i) \leftarrow \mathrm{Beta}(a_i + e_{i,1}(v),b_i + e_{i,0}(v)).
\label{eq:beta-update}
\end{align}
After multiple views, the posterior is
\begin{align}
 p_i \sim \mathrm{Beta}\Big(a_{\mathrm{init}}+ \textstyle\sum_v e_{i,1}(v),\ b_{\mathrm{init}} + \textstyle\sum_v e_{i,0}(v) \Big).
\end{align}
For $a_{\mathrm{init}}{=}b_{\mathrm{init}}$, the posterior mean becomes $a_i/(a_i{+}b_i)$.
In this symmetric case, the Bayes-optimal label reduces to selecting the class with the larger accumulated pseudo-counts, which exactly recovers the decision rule of Eq.~\eqref{eq:fs-solve}:
\begin{align}
 y_i &= \underset{n\in\{0,1\}}{\arg\max} \sum_v e_{i,n}(v) \nonumber\\ 
 &= \underset{n\in\{0,1\}}{\arg\max} \sum_v \sum_{(j,k) \in I(v)} \alpha_i T_i \, \mathbb I[M_{j,k}(v)=n].
\end{align}
Thus, FlashSplat’s label-selection rule appears as the MAP decision within our Bayesian formulation.

\begin{figure}[t]
    \centering
    \includegraphics[width=1.0\columnwidth]{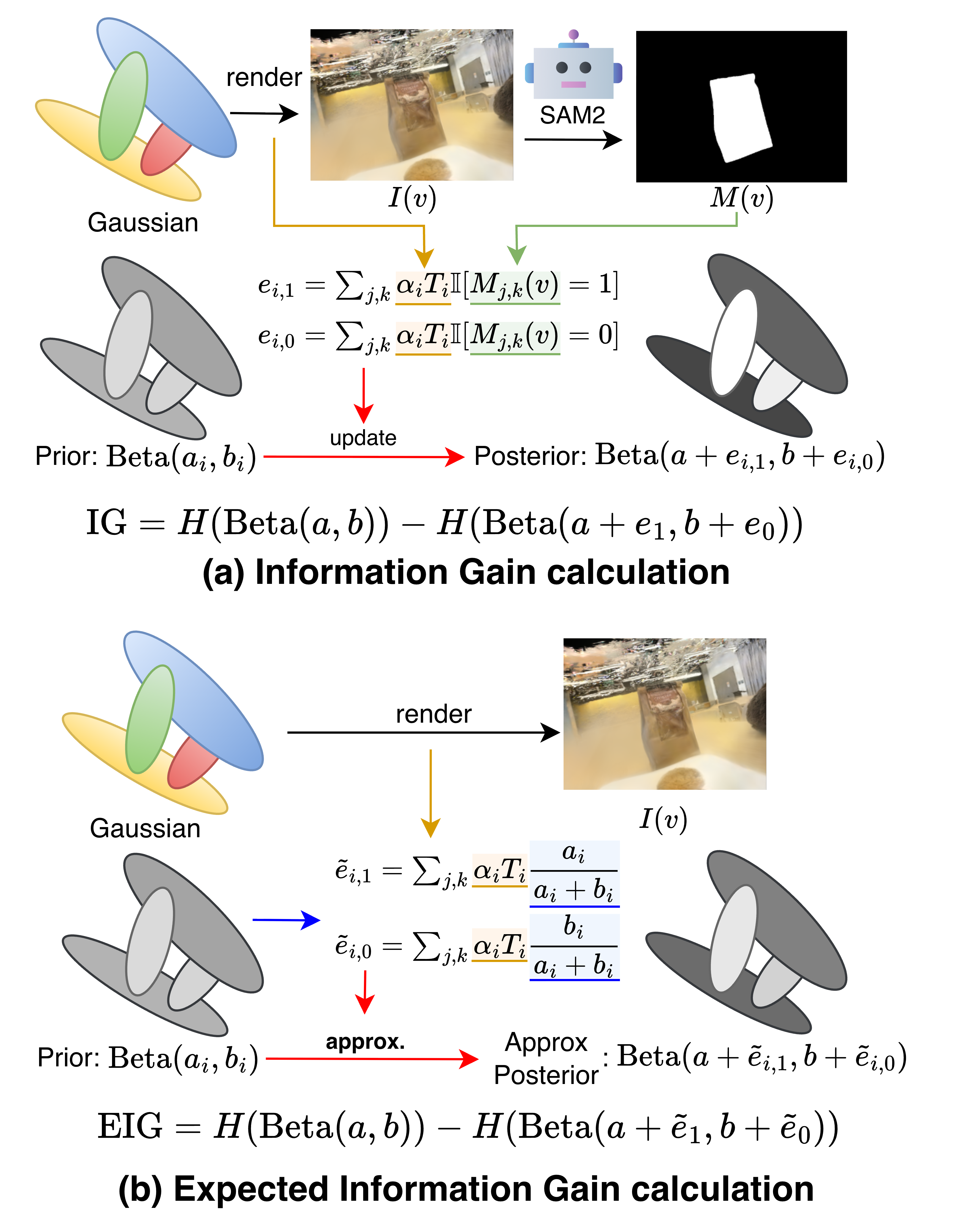} 
\caption{
\textbf{Information Gain vs. Expected Information Gain (ours).}
(a) IG calculation updates the Beta posterior using SAM2 segmentation masks (Eq. \eqref{eq:information-gain}).
(b) Our EIG approximates the posterior update from the prior Beta distribution, avoiding SAM2 inference and enabling efficient viewpoint evaluation (Eq. \eqref{eq:eig}).
}
    \label{fig:eig_explain}
\end{figure}

\subsection{Active Camera Sampling Based on EIG}
\label{sec:eig}
To efficiently estimate $p_i$, we sample the most informative camera view from randomly sampled candidate views. The details of candidate view sampling are described in Sec \ref{sec:overall}.
For each candidate view $v$, the information gain is the difference in entropy before and after adding that view:
\begin{align}
    \mathrm{IG}(v) = &\sum_i \{ H(\mathrm{Beta}(a_i,b_i)) \nonumber\\
    &- H\left(\mathrm{Beta}(a_i+e_{i,1}(v),b_i+e_{i,0}(v))\right)\}
\label{eq:information-gain}
\end{align}
where $H$ denotes entropy and the Beta entropy can be calculated analytically (Appendix~A). 
However, due to the need for mask estimation, it is inefficient to calculate $\mathrm{IG}(v)$ for all candidate views.
Instead, we use \textbf{Expected Information Gain (EIG)}. The per-Gaussian responsibility in the rendered image is calculated as $\tau_i = \sum_{(j,k) \in I(v)} \alpha_i T_i$, and success/failure counts can be approximated using the mean of the Beta distribution $m_i=a_i/(a_i+b_i)$.
\begin{align}
    \tilde e_{i,1} &= m_i \tau_i = \sum_{(j,k) \in I(v)} \alpha_i T_i \frac{a_i}{a_i+b_i} , \nonumber\\
    \tilde e_{i,0} &=(1-m_i)\tau_i  = \sum_{(j,k) \in I(v)} \alpha_i T_i \frac{b_i}{a_i+b_i} . 
    \label{eq:e_tilde}
\end{align}
Assuming $g_i$ falls inside the mask with probability $m_i$, Eq.~\eqref{eq:e_tilde} approximates Eq.~\eqref{eq:information-gain}. 
From them, EIG is calculated as follows:
\begin{align}
    \mathrm{EIG}(v)=&\sum_i \{ H(\mathrm{Beta}(a_i,b_i)) \nonumber\\
    &- H\left(\mathrm{Beta}(a_i+\tilde{e}_{i,1}(v),b_i+\tilde{e}_{i,0}(v))\right)\}
    \label{eq:eig}
\end{align}
Then, we greedily select the view with the highest EIG,
\begin{align}
    v^\star = \arg\max_v \mathrm{EIG}(v)
\end{align}
The stability and validity of EIG-based greedy sampling are discussed in Section~\ref{sec:theory}.

\begin{algorithm}[t]
\caption{B$^3$-Seg: Camera-Free, Training-Free 3DGS Segmentation via Analytic EIG}
\begin{algorithmic}[1]
\STATE Initialize $(a_i,b_i)\leftarrow(a_\mathrm{init},b_\mathrm{init})$ for all Gaussians.
\STATE From a canonical view of the scene, estimate an initial mask using segmentation module of Sec. \ref{sec:openvocab}.
\STATE Update $(a_i,b_i)$ with Eq.~\eqref{eq:beta-update}.
\STATE Compute $\mathbf{c}_{\mathrm{obj}}, r_{\mathrm{obj}}$ from Gaussians satisfying $a_i > b_i$.
\FOR{iteration $t=1,\dots,T$}
  \STATE Uniformly sample $N_{\mathrm{cand}}$ candidate viewpoints on the sphere centered at $\mathbf{c}_{\mathrm{obj}}$, with a radius $r_\mathrm{sphere}$.
    \FOR{each candidate $v$}
      \STATE Render once to obtain $\tau_i = \sum_{(j,k)} \alpha_i T_i$.
      \STATE Compute $(\tilde{e}_{i,1}, \tilde{e}_{i,0})$ with Eq.~\eqref{eq:e_tilde}.
      \STATE Compute $\mathrm{EIG}(v)$ with Eq.~\eqref{eq:eig}.
    \ENDFOR
    \STATE Select $v^\star = \arg\max_v \mathrm{EIG}(v)$.
  \STATE Render RGB image $I(v^\star)$.
  \STATE Estimate $M(v^\star)$ with the module of Sec. \ref{sec:openvocab}.
  \STATE Compute $(e_{i,1}(v^\star), e_{i,0}(v^\star))$ with Eq.~\eqref{eq:agg-evidence}.
  \STATE Beta update: $(a_i,b_i)\!\leftarrow\!(a_i+e_{i,1},\,b_i+e_{i,0})$.
  \STATE Update $\mathbf{c}_{\mathrm{obj}}, r_{\mathrm{obj}}$ from Gaussians with $a_i > b_i$.
\ENDFOR
\STATE Return final 3D mask: Gaussians with $a_i > b_i$ 
\end{algorithmic}
\end{algorithm}

\subsection{Open-Vocabulary Mask Inference}
\label{sec:openvocab}

To obtain a 2D semantic mask for each selected view, we adopt a lightweight open-vocabulary segmentation module. 
This module consists of three stages:

\begin{enumerate}
\item \textbf{Text-conditioned region proposal (Grounding DINO).}
Given a user prompt, Grounding DINO~\cite{grounding-dino} generates
candidate bounding boxes $\{B_k\}$ that indicate possible object regions.

\item \textbf{Mask prediction with prior guidance (SAM2).}
For each candidate box $B_k$ in a view $v^\star$, we use SAM2~\cite{sam2} for segmentation masks.
To stabilize inference, SAM2's \texttt{mask\_input} receives a prior image $R(v^\star)$, rendered from current Beta means $m_i=\tfrac{a_i}{a_i+b_i}$: 
\begin{align*}
    R_{\mathrm{soft}}(v^\star) = \sum_i m_i\,\alpha_i\,T_i(v^\star), ~R(v^\star) = \log\frac{R_{\mathrm{soft}}(v^\star)}{1-R_{\mathrm{soft}}(v^\star)}.
\end{align*}

This cues SAM2 with view-specific information from Beta posterior means, ensuring temporal consistency and reducing drift to distractors.

\begin{align*}
    M_k(v^\star) = \mathrm{SAM2}\left(I(v^\star),\; B_k,\; \texttt{mask\_input}=R(v^\star)\right).
\end{align*}

\item \textbf{Semantic ranking via CLIP.}
Each candidate mask $M_k(v)$ is applied to the RGB image to obtain a masked crop,
which is scored by CLIP~\cite{clip} against the user text. The highest scoring mask for the view $v$ is chosen:
\begin{align*}
    M(v^\star) = \arg\max_{k} \mathrm{CLIP}\big(I(v^\star) \odot M_k(v^\star),\; \text{text}\big).
\end{align*}
\end{enumerate}
Following the selected mask, $e_{i,1}(v^\star)$ and $ e_{i,0}(v^\star)$ are calculated with Eq.~\eqref{eq:agg-evidence}.
We then update the Beta parameters according to Eq. ~\eqref{eq:beta-update}:
\begin{align*}
\mathrm{Beta}(a_i, b_i) \leftarrow \mathrm{Beta}(a_i + e_{i,1}(v^\star), b_i + e_{i,0}(v^\star)).
\end{align*}
We repeat this view-selection and update loop.

\subsection{Overall Pipeline and Candidate Sampling}
\label{sec:overall}
Algorithm~1 summarizes the pipeline. 
We initialize all Gaussians with $(a_{\mathrm{init}}, b_{\mathrm{init}})$ and obtain an initial mask from a canonical view using the procedure in Section~\ref{sec:openvocab}. 
After this first update, we consider Gaussians with $a_i>b_i$ as foreground and estimate the object center and radius by 
\begin{align*}
\mathbf{c}_{\mathrm{obj}} = \frac{\sum_{i\in\mathcal{G}_{\mathrm{fg}}} m_i\,\mu_i}{\sum_{i\in\mathcal{G}_{\mathrm{fg}}} m_i}, 
~r_{\mathrm{obj}} = \frac{\sum_{i\in\mathcal{G}_{\mathrm{fg}}} m_i \,\|\mu_i - \mathbf{c}_{\mathrm{obj}}\|}{\sum_{i\in\mathcal{G}_{\mathrm{fg}}} m_i}.
\end{align*}
Then, $N_{\mathrm{cand}}$ candidate cameras are uniformly sampled from the $\mathbf{c}_{\mathrm{obj}}$ centered sphere, with a radius calculated as $r_{\mathrm{sphere}} = 1.5~r_{\mathrm{obj}} / \tan(\mathrm{fov}/2)$.
Next, we compute the EIG for each candidate and pick the most informative view.
We repeat mask inference and updates $(a_i,b_i)$, while updating $\mathbf{c}_{\mathrm{obj}}, r_{\mathrm{obj}}$ at each iteration. 
The final 3D mask consists of Gaussians with $a_i>b_i$.

\begin{figure*}[t]
    \centering
    \includegraphics[width=0.89\linewidth]{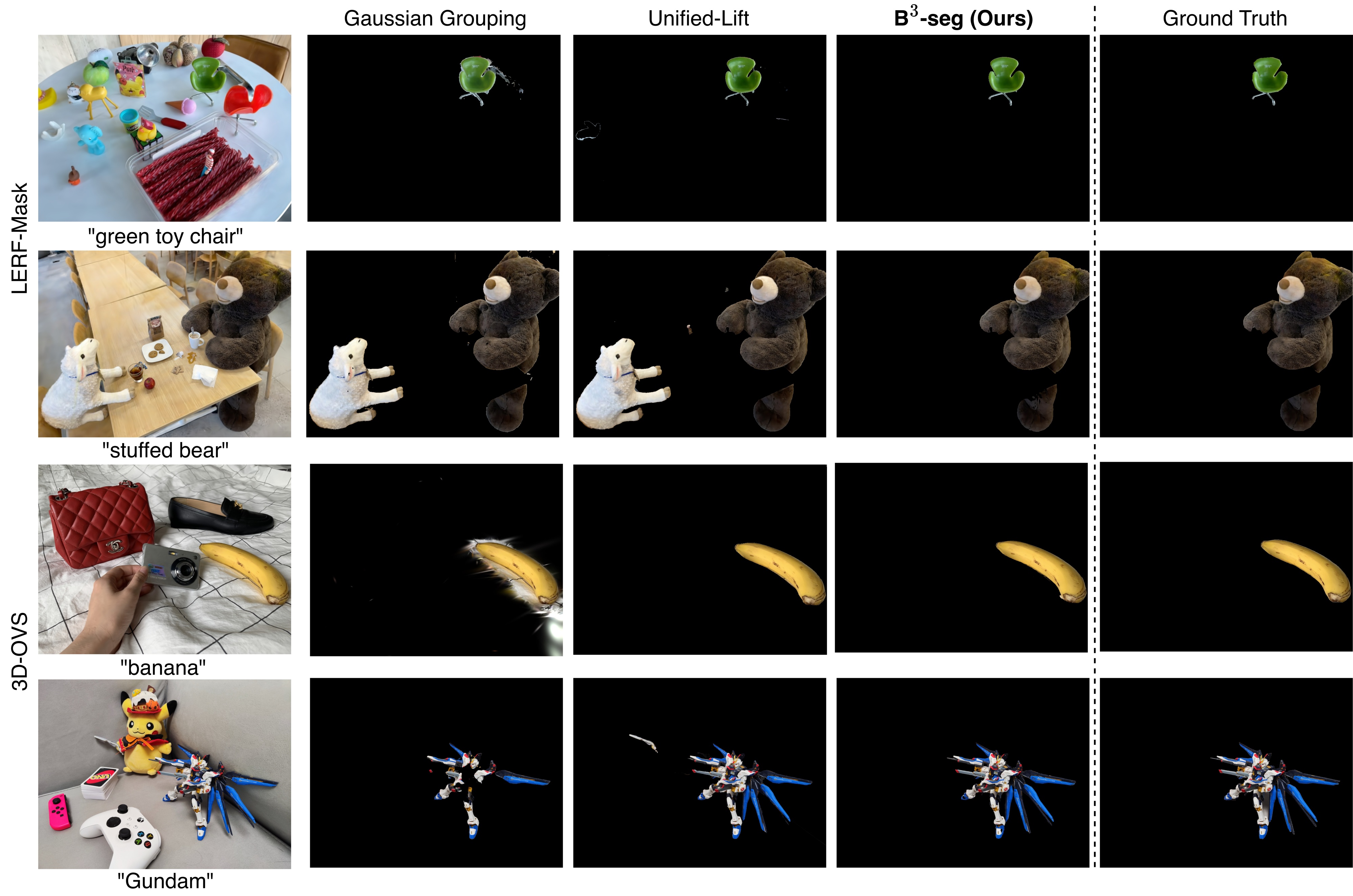} 
    \caption{Qualitative comparison on text-guided 3D segmentation.
We compare our method (\textbf{B$^{3}$-Seg}) with prior 3DGS segmentation approaches.
Our method produces cleaner and more complete object masks, especially in cluttered scenes.}
    \label{fig:qualitative}
\end{figure*}

\begin{figure*}[t]
    \centering
    \includegraphics[width=0.85\linewidth]{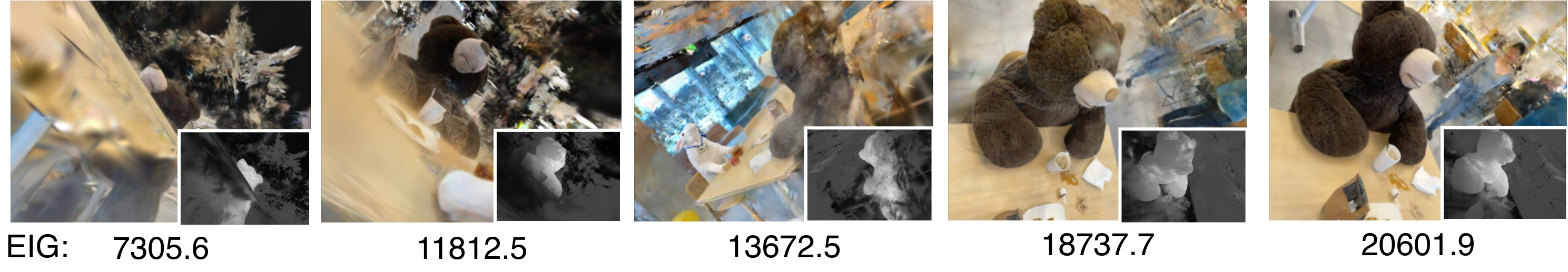} 
    \caption{Candidate-view EIG on LERF-Mask (Teatime) with the prompt \emph{``stuffed bear''}. Each panel shows a candidate rendering; the bottom-right inset is the current confidence map (posterior mean).}
    \label{fig:eig_comparison}
\end{figure*}

\section{Theoretical Guarantees}
\label{sec:theory}

This section shows that the proposed method is (i) adaptive monotone and (ii) adaptive submodular, and then derives (iii) a greedy $(1{-}1/e)$ approximation guarantee. 
Let $S$ be the set of already selected views, and denote by $\mathrm{EIG}(v\mid S)$ the conditional expected gain when adding a candidate view $v$ under $S$ (the per-view EIG itself is given in Eq.~\eqref{eq:eig}).
The detailed proofs are provided in the Appendix.

\medskip\noindent\textbf{Lemma 1 (Adaptive monotonicity).}
$\mathrm{EIG}(v\mid S)$ is adaptive monotone, i.e., $\mathbb{E}[\mathrm{EIG}(v\mid S)]\ge0$ for any selected set $S$ and any candidate $v$.\\
\emph{Sketch.} The Beta entropy $H(\mathrm{Beta}(a_i,b_i))$ is non-increasing in $\kappa_i=a_i{+}b_i$ for $\kappa_i \ge 2$ (Appendix.~\ref{app:proof-nonneg}), which is satisfied by our initialization ($a_{\mathrm{init}}=b_{\mathrm{init}}=1$).
Each candidate $v$ adds nonnegative pseudo-counts $\tau_i(v)\ge0$ to $\kappa_i$, so the posterior entropy decreases in expectation, which yields $\mathbb{E}[\mathrm{EIG}(v\mid S)]\ge0$. 
Intuitively, observing an additional view always decreases uncertainty, so the learning progress is stable.

\medskip\noindent\textbf{Lemma 2 (Adaptive submodularity).}
$\mathrm{EIG}(v\mid S)$ is adaptive submodular, i.e., $\mathbb{E}[\mathrm{EIG}(v\mid S)]\ge \mathbb{E}[\mathrm{EIG}(v\mid S')]$ for any $S'\supseteq S$ and any candidate $v$. \\
\emph{Sketch.} As we observe more views, each $\kappa_i$ increases and the reduction in Beta entropy (Eq.~\eqref{eq:eig}) becomes smaller (Appendix.~\ref{app:proof-submod}). 
In addition, $\tau_i(v)$ is added linearly and is nonnegative. 
Therefore, the expected marginal gain for the same $v$ is smaller under the larger set $S'$. 

\medskip\noindent\textbf{Theorem (Greedy $(1{-}1/e)$ approximation).}
At each step $t$, we greedily select 
$v_t^{\text{greedy}} = \arg\max_v\, \mathrm{EIG}(v \mid S_{t-1})$,
where $S_{t-1}$ is the set of views already selected.
We define the total expected information gain of this greedy policy as
\[
\mathrm{EIG}(S_k^{\text{greedy}}) := \sum_{t=1}^{k}\, \mathrm{EIG}(v_t^{\text{greedy}} \mid S_{t-1}^{\text{greedy}}).
\]
Then, according to Theorem~16 of~\cite{golovin2011adaptive},
\begin{align}
\mathbb{E}\!\left[\mathrm{EIG}(S_k^{\text{greedy}})\right]
\;\ge\;
(1 - 1/e)\,
\max_{\pi}\, \mathbb{E}\!\left[\mathrm{EIG}(S_k^{\pi})\right],
\end{align}
where $S_k^{\pi}$ denotes the sets obtained by all adaptive policies. Thus, our EIG-based greedy selection achieves a $(1{-}1/e)$ approximation to the optimal view selection policy.

\section{Experiment}
\subsection{Experimental Settings}
We evaluated on two datasets: LERF-Mask \cite{gaussian-grouping} and 3D-OVS \cite{3dovs}. 
For LERF-Mask, we follow the Gaussian Grouping protocol \cite{gaussian-grouping}; for 3D-OVS, we follow \cite{objectgs}. 
We set the number of update iterations to $T{=}20$, the number of candidate views to $N_{\mathrm{cand}}{=}20$, and initialize the Beta parameters with $a_{\mathrm{init}}{=}b_{\mathrm{init}}{=}1$. 
Each sequence starts with a single canonical scene view, providing the initial mask and object center. The canonical view is the first camera viewpoint in the scene metadata, which does not need extra supervision or specific reconstruction data—just the camera pose of dataset. It is used only to obtain the initial prior, with subsequent views chosen by our EIG-based planning. The low dependency on the initial condition is discussed in Sec. \ref{sec:ablation}
Our experiments were conducted with a single RTX A6000 GPU. 
The end-to-end runtime (rendering, mask inference, and updates) is within a few seconds, enabled by our approximation and active selection (Section~\ref{sec:theory}).

\begin{table*}[t]
\centering
\setlength{\tabcolsep}{5.2pt}
\begin{tabular}{@{}lccccccc@{}}
\toprule
\multirow{2}{*}{Method} &
\multicolumn{4}{c}{Accuracy (mIoU / mBIoU)} &
\multirow{2}{*}{views} &
\multirow{2}{*}{time} &  
\multirow{2}{*}{steps} \\
\cmidrule(lr){2-5}
& figurines & ramen & teatime & mean & & & \\ 
\midrule
\multicolumn{7}{l}{\textit{Assumes reconstruction views and/or labels (not directly comparable)}}\\
\addlinespace[2pt]
LERF~\cite{lerf} & 33.5 / 30.6 & 28.3 / 14.7 & 49.7 / 42.6 & 37.2 / 29.3 & GT & 45 min & 30k\\
SA3D~\cite{sa3d} & 24.9 / 23.8 & 7.4 / 7.0 & 42.5 / 39.2 & 24.9 / 23.3 & GT & 35 min & 30k\\
LangSplat~\cite{langsplat} & 52.8 / 50.5 & 50.4 / 44.7 & 69.5 / 65.6 & 57.6 / 53.6 & GT & 19 min & 30k\\
Gaussian Grouping~\cite{gaussian-grouping} & 69.7 / 67.9 & 77.0 / 68.8 & 71.7 / 66.1 & 72.8 / 67.6 & GT & 37 min & 30k\\
Gaga~\cite{gaga} & \textbf{90.7 / 89.0} & 64.1 / 61.6 & 69.3 / 66.0 & 74.7 / 72.2 & GT & 13 min & 30k\\
Unified-Lift~\cite{unifiedlift} & -- & -- & -- & 80.9 / 77.1 & GT & 40 min & 30k \\
ObjectGS~\cite{objectgs} & 88.2 / 89.0 & \textbf{88.0 / 79.9} & \textbf{88.9 / 88.6} & \textbf{88.4 / 85.8} & GT & $\sim$ 50 min & 30k\\
\midrule
\multicolumn{7}{l}{\textit{Sampling-based, no retraining (directly comparable within this block)}}\\
\addlinespace[2pt]
FlashSplat \cite{flashsplat} (Uniform\mbox{-}Sphere)$^{\dagger}$ & 60.2 / 57.5 & 68.4 / 61.5 & 80.4 / 76.3 & 69.6 / 65.1 & Sample & \textbf{10.2 sec} & \textbf{20}\\
FlashSplat \cite{flashsplat} (Recon\mbox{-}Cam)$^{\ddagger}$      & 71.6 / 69.1 & 71.4 / 66.3 & 86.6 / 83.9 & 76.5 / 73.1 & Sample & \textbf{10.1 sec} & \textbf{20}\\
\textbf{B$^3$-Seg (Ours)}                        & \textbf{88.3 / 85.4} & \textbf{75.3 / 69.7} & \textbf{89.8 / 88.0} & \textbf{84.5 / 81.0} & \textbf{Sample} & \textbf{12.1 sec} & \textbf{20}\\
\bottomrule
\end{tabular}
\caption{\textbf{LERF-Mask (accuracy, assumptions, and latency).}
Top: Methods that require reconstruction views/labels (=not directly comparable).
Bottom: Sampling-based, training-free approach with our 20 views/updates runtime (few seconds). $^{\dagger}$ \textit{Uniform-Sphere}: Candidate viewpoints sampled uniformly on a sphere. $^{\ddagger}$ \textit{Recon-Cam}: Candidate viewpoints randomly sampled from reconstruction cameras.}
\label{tab:lerf-mask}
\end{table*}

\subsection{Qualitative Results}
Figure~\ref{fig:qualitative} shows representative comparisons in the LERF-Mask and 3D-OVS dataset. 
Our B$^{3}$-Seg produces cleaner, more complete object masks than prior 3DGS methods; for example, it recovers the full chair in the ``green toy chair'' scene and segments the ``stuffed bear'' despite heavy clutter. 
These improvements follow from two complementary components: analytic EIG that prioritizes highly informative views, and robust Beta-Bernoulli updates that accumulate reliable pseudo-counts from the 2D mask. 
Importantly, almost high-mIoU baselines rely on reconstruction views, ground-truth masks, or per-scene optimization that take tens of minutes. 
By contrast, our B$^{3}$-Seg achieves comparable visual quality in a budget of 20 views and a few seconds.

Figure~\ref{fig:eig_comparison} illustrates our view selection on the LERF-Mask Teatime scene. 
Views where the bear is larger and less occluded have a higher EIG, showing that our view-selection criterion selects views that reduce uncertainty.

\begin{table}[t]
\centering
\setlength{\tabcolsep}{5pt}
\begin{tabular}{@{}lcccc@{}}
\toprule
\multirow{2}{*}{Method} & \multicolumn{4}{c}{mIoU (\%)} \\
\cmidrule(lr){2-5}
 & Bed & Bench & Sofa & Lawn \\
\midrule
\multicolumn{5}{l}{\textit{Assumes reconstruction views/labels (not comparable)}}\\
\addlinespace[2pt]
LangSplat~\cite{langsplat}            & 92.5 & 94.2 & 90.0 & 96.1 \\
Feature 3DGS~\cite{featuregs}      & 83.5 & 90.7 & 86.9 & 93.4 \\
LEGaussians~\cite{legaussians}        & 84.9 & 91.1 & 87.8 & 92.5 \\
Gaussian Grouping~\cite{gaussian-grouping} & 83.0 & 91.5 & 87.3 & 90.6 \\
N2F2~\cite{n2f2}                      & 93.8 & 92.6 & 92.1 & 96.3 \\
SAGA~\cite{saga}                      & 97.4 & 95.4 & 93.5 & \textbf{96.6} \\
FastLGS~\cite{fastlgs}                & 94.7 & 95.1 & 90.6 & 96.2 \\
LBG~\cite{lbg}                        & 97.7 & 96.3 & \textbf{97.3} & 87.4 \\
CCL-LGS~\cite{ccl-lgs}                & 97.3 & 95.0 & 92.3 & 96.1 \\
ObjectGS~\cite{objectgs}              & \textbf{98.0} & \textbf{96.4} & 97.2 & 95.4 \\
\midrule
\multicolumn{5}{l}{\textit{Camera-free \& training-free (comparable)}}\\
\addlinespace[2pt]
FlashSplat (Uniform\mbox{-}Sphere) & 91.7 & 86.9 & 90.2 & 91.9 \\
FlashSplat (Recon\mbox{-}Cam)    & 94.3 & 90.3 & 85.7 & 96.3 \\
\textbf{B$^3$-Seg (Ours)}                     & \textbf{97.1} & \textbf{92.2} & \textbf{94.1} & \textbf{96.8} \\
\bottomrule
\end{tabular}
\caption{\textbf{3D\mbox{-}OVS (mIoU).} Top: assumes views/labels (\emph{not comparable}). Bottom: camera-free \& training-free.}
\label{tab:3d-ovs}
\end{table}

\subsection{Quantitative Results}

\noindent\textbf{LERF-Mask.}\quad Table~\ref{tab:lerf-mask} reports accuracy and latency. 
As baselines, FlashSplat (Uniform-Sphere) and FlashSplat (Recon-Cam) are used to isolate sampling effects.
In FlashSplat (Uniform-Sphere), $N_\mathrm{cand}$ candidate viewpoints are sampled uniformly on a sphere centered at the estimated object center; a random candidate chosen (no camera priors, no EIG). 
In FlashSplat (Recon-Cam),  $N_\mathrm{cand}$ candidate viewpoints are randomly sampled from reconstruction cameras (available on LERF-Mask).
Both baselines used the same segmentation pipeline with B$^3$-seg for a fair comparison.
From Table~\ref{tab:lerf-mask}, B$^3$-Seg achieves higher segmentation scores than baselines and is competitive even with methods that assume reconstruction views.

\noindent\textbf{3D-OVS.}\quad Table~\ref{tab:3d-ovs} follows the same presentation: the top block lists methods that assume reconstruction views/labels, while the bottom block compares sampling-based, training-free approaches under the 20-view budget. 
As shown in Table~\ref{tab:3d-ovs}, our B$^3$-Seg surpasses baselines in segmentation scores and matches the performance of methods with reconstruction views.

\begin{figure}[t]
    \centering
    \includegraphics[width=0.95\columnwidth]{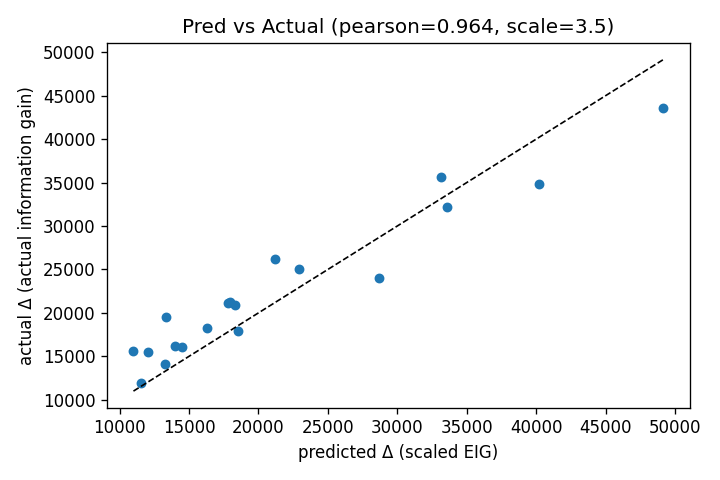} 
\caption{Predicted EIG closely matches information gain on the LERF-Mask, with a strong correlation ($r=0.964$).}
    \label{fig:scatter_pred_vs_actual}
\end{figure}

\noindent\textbf{EIG proxy validation.}\quad We further verify that our analytic EIG is a faithful proxy for the true information gain. 
Figure~\ref{fig:scatter_pred_vs_actual} plots the analytic EIG (Eq.~\eqref{eq:eig}) against the measured information gain (Eq.~\eqref{eq:information-gain}) for selected views in scenes. 
A strong correlation of $\mathrm{IG}(v)$ and $\mathrm{EIG}(v)$ supports the use of Eq.~\eqref{eq:eig} as a practical ranking surrogate.

\begin{table}[t]
\centering
\setlength{\tabcolsep}{6pt}
\begin{tabular}{@{}ccc ccc@{}}
\toprule
\multirow{2}{*}{\shortstack{CLIP\\re-rank}} &
\multirow{2}{*}{\shortstack{SAM2\\mask-input}} &
\multicolumn{2}{c}{Accuracy} & \multicolumn{2}{c}{$\Delta$ vs.~Base} \\
\cmidrule(lr){3-4} \cmidrule(lr){5-6}
& & mIoU & mBIoU & mIoU & mBIoU \\
\midrule
--          & --          & 74.9 & 70.2 & --   & --   \\
\checkmark  & --          & 81.5 & 76.6 & +6.6 & +6.4 \\
\checkmark  & \checkmark  & 84.5 & 81.0 & +9.6 & +10.8 \\
\bottomrule
\end{tabular}
\caption{\textbf{Ablation study on LERF-Mask.} Both CLIP re-ranking and SAM2 mask-input improve segmentation performance.}
\label{tab:ablation}
\end{table}

\begin{figure}[t]
    \centering
    \includegraphics[width=0.9\columnwidth]{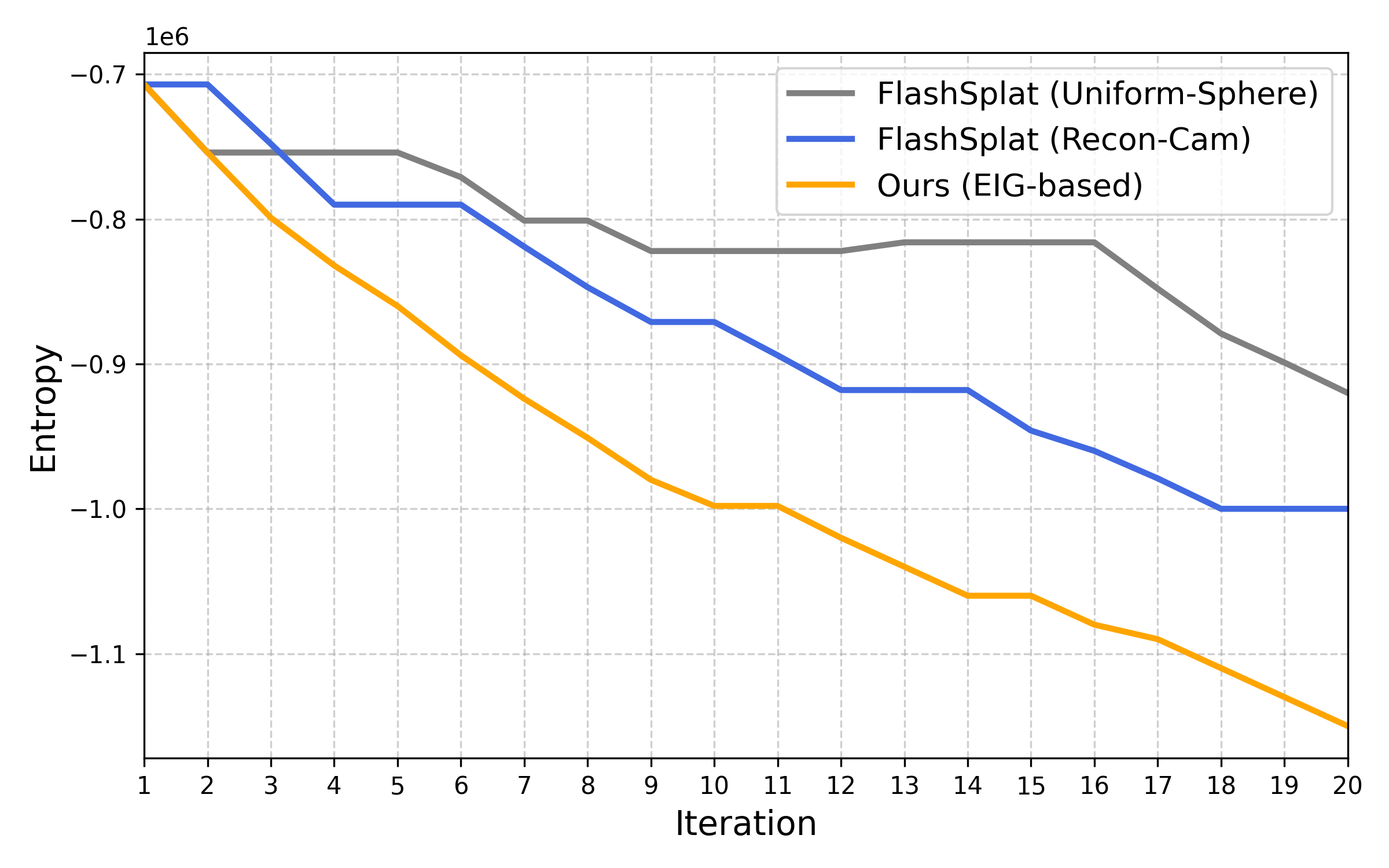} 
\caption{Posterior entropy vs. iteration for three view-selection strategies on the LERF-Mask Teatime scene. EIG-based selection consistently achieves the largest per-step entropy drop.}
    \label{fig:entropy_plot}
\end{figure}

\noindent\textbf{Information efficiency over iterations.}\quad Figure~\ref{fig:entropy_plot} visualizes the total Beta entropy $\sum_i H(\mathrm{Beta}(a_i, b_i))$ in iterations on the LERF-Mask Teatime scene. 
Lower entropy indicates higher segmentation certainty. 
Our EIG-driven policy achieves the largest per-step entropy drop and converges faster than spherical and reconstruction-view sampling, confirming improved information-efficiency.

\noindent\textbf{Runtime analysis.}\quad 
Table~\ref{tab:runtime_breakdown} reports the runtime breakdown of our method.  
Mask inference is the dominant cost (9.76\,s out of 12.11\,s). Our EIG-based view selection only requires lightweight rendering and entropy evaluation, avoiding mask inference on each candidate view.

\subsection{Ablation Studies}
\label{sec:ablation}

\noindent\textbf{CLIP re-ranking and SAM2 mask-input}
We evaluated the effectiveness of (i) CLIP re-ranking of candidate masks and (ii) SAM2 with prior mask input.
Integrating these improves segmentation quality (Table~\ref{tab:ablation}). CLIP filters out inconsistent candidates and SAM2 stabilizes masks. 
These enhancements complement the EIG-based selection. 

\noindent\textbf{Sensitivity analysis.}
We evaluate the influence of $N_{\mathrm{cand}}$ and $T$ in Table~\ref{tab:sensitivity_joint}. 
Accuracy improves as these parameters increase but quickly saturates around $N_{\mathrm{cand}}{=}20$ and $T{=}20$, showing that our lightweight setting is already sufficient.
Furthermore, as shown in Table~\ref{tab:init_center_sensitivity}, B$^{3}$-Seg remains robust even when the initial object center is perturbed: a 50\% shift in $\mathbf{c}_{\mathrm{obj}}$ results in only a 1.6\% drop in mIoU.
This illustrates that the performance of B$^3$-Seg is not dependent on its initial condition; EIG-based active view selection quickly compensates for any suboptimal initial condition.

\begin{table}[t]
\centering
\setlength{\tabcolsep}{4pt}
\begin{tabular}{lcccc|c}
\toprule
 & \begin{tabular}[c]{@{}c@{}}Mask\\Inference\end{tabular}
 & \begin{tabular}[c]{@{}c@{}}View\\Selection\end{tabular}
 & \begin{tabular}[c]{@{}c@{}}Beta\\Update\end{tabular}
 & \begin{tabular}[c]{@{}c@{}}Other \end{tabular}
 & Total \\
\midrule
Time (s) & 9.761 & 2.111 & 0.119 & 0.117 & 12.108 \\
\bottomrule
\end{tabular}
\caption{\textbf{Runtime breakdown of B$^3$-Seg}.
End-to-end inference completes in 12 seconds for 20 actively selected views.}
\label{tab:runtime_breakdown}
\end{table}

\begin{table}[t]
\centering
\setlength{\tabcolsep}{6pt}
\renewcommand{\arraystretch}{1.2}

\begin{tabular}{@{}lcccc@{}}
\toprule
$N_{\mathrm{cand}}$ & 5 & 10 & 20 & 30 \\
\midrule
mIoU (\%)        &  76.2  &  83.1  &  84.5  &  84.6  \\
time (s)         &  10.3  &  11.6  &  12.1  &  14.4  \\
\midrule
$T$ (iterations) &  5 & 10 & 20 & 30 \\
\midrule
mIoU (\%)        &  79.8  &  82.7  &  84.5  &  84.8  \\
time (s)         &  3.5   &  6.6   &  12.1  &  19.4  \\
\bottomrule
\end{tabular}

\caption{\textbf{Sensitivity to $N_{\mathrm{cand}}$ and $T$ on LERF-Mask.}
Accuracy saturates around $N_{\mathrm{cand}}{=}20$ and $T{=}20$ while runtime increases almost linearly.}
\label{tab:sensitivity_joint}
\end{table}

\begin{table}[t]
\centering
\setlength{\tabcolsep}{6pt}
\renewcommand{\arraystretch}{1.2}

\begin{tabular}{@{}lccccc@{}}
\toprule
$\mathbf{c}_{\mathrm{obj}}$ shift & 0\% & 20\% & 50\% & 70\% & 100\% \\
\midrule
mIoU (\%) & 84.5 & 83.6 & 82.9 & 81.7 & 80.7 \\
$\Delta$  & --   & -0.9 & -1.6 & -2.8 & -3.8 \\
\bottomrule
\end{tabular}

\caption{\textbf{Sensitivity to the initial condition on LERF-Mask.}
We shift the initial object center $\mathbf{c}_{\mathrm{obj}}$ along a random 3D direction by a percentage of $r_{\mathrm{obj}}$.}
\label{tab:init_center_sensitivity}
\end{table}

\subsection{Discussion and Limitations}
\label{sec:limitations}

B$^3$-Seg offers a fast and theoretically grounded framework for camera-free 3DGS segmentation.
Our experiments focus on typical object-centric scenes, where the proposed analytic EIG and sequential Bayesian updates operate efficiently. Extending the method to substantially larger
environments—such as wide indoor spaces or outdoor scans—may require broader viewpoint exploration strategies. Integrating methods like RRT-based camera sampling or multi-scale candidate generation
is a promising direction that remains compatible with our analytic EIG formulation.

Additionally, the present probabilistic model addresses binary foreground–background decisions.
Many real-world applications involve multiple objects or semantic categories. 
The framework naturally generalizes to a Dirichlet–Categorical model. This enables multi-object segmentation while preserving the pseudo-count interpretation and the theoretical properties of our EIG-driven updates.

\section{Conclusion}
We introduce B$^3$-Seg, a camera-free, training-free, open-vocabulary framework for 3DGS segmentation. The method uses sequential Beta--Bernoulli updates for per-Gaussian labeling and selects camera views via analytic EIG. This Bayesian approach ensures adaptive monotonicity and adaptive submodularity, providing a \emph{$(1{-}1/e)$} approximation guarantee. B$^3$-Seg achieves competitive scores with the latest methods, without using camera trajectories or ground-truth labels.
Our Bayesian framework can be generalized to multi-class segmentation with a Dirichlet--Categorical model and scalability for larger or dynamic scenes, all integrable into the current EIG-based pipeline. These are left for future work.

\newpage

{
    \small
    \bibliographystyle{ieeenat_fullname}
    \bibliography{main}
}


\clearpage
\newpage

\setcounter{page}{1}
\maketitlesupplementary

\appendix
\section{Beta Entropy and Notation}
\label{app:beta-entropy}
Let $X\sim\mathrm{Beta}(a,b)$ with $a,b>0$. 
For simplicity, we define $B(\cdot,\cdot)$ as a Beta function, and $H(\cdot,\cdot)$ as the entropy of the Beta distribution.
The entropy is
\begin{align}
H(a,b)
=& \log B(a,b) - (a-1)\,\psi(a) - (b-1)\,\psi(b)\nonumber\\
&+ (a+b-2)\,\psi(a+b),
\label{eq:beta-entropy-ja}
\end{align}
where $\psi$ is the digamma function. 

\section{Lemma 1: Non-negativity (Adaptive Monotonicity)}
\label{app:proof-nonneg}
\textbf{Claim.} For any candidate view $v$ and any selected set $S$, the expected gain is nonnegative: $\mathbb{E}[\mathrm{EIG}(v\mid S)]\ge 0$.

\noindent\textbf{Proof.} 
For a single Gaussian $i$, define its mean $m_i$ and concentration $\kappa_i$
\[
    m_i = \frac{a_i}{a_i + b_i}, \qquad
    \kappa_i = a_i + b_i.
\]

From Eq. \eqref{eq:eig}, the approximate posterior entropy of EIG is given by
\begin{align*}
    &H(a_i+\tilde{e}_{i,1}(v),b_i+\tilde{e}_{i,0}(v))\\
    &=H(a_i+m_i\tau_i, b_i+(1-m_i)\tau_i)\\
    &=H\Bigg(\bigg(1 + \sum_{(j,k) \in I(v)} \alpha_i T_i \frac{1}{a_i+b_i}\bigg)a_i, \\
    &\qquad \qquad\bigg(1 + \sum_{(j,k) \in I(v)} \alpha_i T_i \frac{1}{a_i+b_i}\bigg)b_i\Bigg)\\
    &:=H(c_ia_i, c_ib_i).
\end{align*}

This corresponds to multiplying the concentration $\kappa_i$ by $c_i>1$ while leaving the mean $m_i$ unchanged.
This can be interpreted as the observation of more trials of the same underlying Bernoulli probabilities.

In this situation, increasing the concentration $\kappa_i$ makes the Beta distribution more peaked around $m_i$ and therefore reduces its entropy.
In particular, $\mathrm{Beta}(1,1)$ is the uniform distribution, which has the maximum entropy among Beta distributions.
In our setting of $a_i,b_i>1$, the density becomes increasingly concentrated
and the entropy decreases.

Equivalently,  we obtain
\[
    H(c_i a_i, c_i b_i) \;\le\; H(a_i,b_i)
\]
for every Gaussian $i$.
Thus the entropy drop
\[
    \Delta H_i(v\mid S)
    := H_i(S)
     - H_i\bigl(a_i(S)+e_{i,1}(v),\, b_i(S)+e_{i,0}(v)\bigr)
\]
is nonnegative, and summing over $i$ yields
\(
    \mathrm{EIG}(v\mid S) \ge 0.
\)
Taking expectation over the randomness of the observation preserves the inequality, so we have
\[
    \mathbb{E}[\mathrm{EIG}(v\mid S)] \ge 0.
\]

\section{Lemma 2: Adaptive Submodularity}
\label{app:proof-submod}

\noindent\textbf{Claim.}
For any $S\subseteq S'$ and any candidate view $v$,
\[
    \mathbb{E}[\mathrm{EIG}(v\mid S)]
    \;\ge\;
    \mathbb{E}[\mathrm{EIG}(v\mid S')].
\]

\noindent\textbf{Proof.}
In our formulation, the uncertainty of each Gaussian $i$ is represented by
the entropy of its Beta posterior
\[
    H_i(S) = H(a_i(S),\,b_i(S)).
\]
The per-view information gain is defined as the total entropy reduction:
\[
    \mathrm{EIG}(v\mid S)
    =
    \sum_i \Bigl[
        H_i(S)
        - H_i(a_i(S)+e_{i,1}(v),\; b_i(S)+e_{i,0}(v))
    \Bigr],
\]
which is identical to Eq. \eqref{eq:eig} in the main paper.

The success/failure count increment
\[
    \tau_i(v) = e_{i,1}(v)+e_{i,0}(v)
\]
depends only on the chosen view $v$ and is independent of $S$.
On the other hand, if $S\subseteq S'$, then
\[
    \kappa_i(S)=a_i(S)+b_i(S)
    \;\le\;
    \kappa_i(S')=a_i(S')+b_i(S'),
\]
meaning that the concentration monotonically increases as more views are observed.

By Lemma 1, the Beta entropy $H_i(\kappa_i)$ is a monotonically decreasing
function of the concentration parameter $\kappa_i$ (with the mean fixed).  
In other words, as the pseudo-counts accumulate and $\kappa_i$ becomes larger,
the posterior becomes more concentrated and its entropy becomes less sensitive
to additional evidence.  
Therefore, adding the same pseudo-count increment $\tau_i(v)$ results in a
smaller change in entropy when $\kappa_i$ is large. 
When $\kappa_i$ is small, the posterior
distribution is still broad, so new evidence can significantly reduce
uncertainty.    
In contrast, when $\kappa_i$ is already large and the posterior is sharply
concentrated, the same $\tau_i(v)$ leads to only a minor decrease in uncertainty.
Therefore, when $S\subseteq S'$,
\begin{align}
    \Delta H_i(\kappa_i(S);\tau_i(v))
    \;\ge\;
    \Delta H_i(\kappa_i(S');\tau_i(v)).
    \label{eq:submod}
\end{align}

Summing Eq. \eqref{eq:submod} over all Gaussians $i$, and taking expectation over the
random observation, yields
\[
    \mathbb{E}[\mathrm{EIG}(v\mid S)]
    \;\ge\;
    \mathbb{E}[\mathrm{EIG}(v\mid S')],
\]
which proves the adaptive submodularity of the proposed EIG.

Intuitively, as more views are incorporated into $S'$, the concentration
$\kappa_i$ of each Gaussian increases and the Beta posterior becomes more
peaked.
Additional observations therefore reduce uncertainty by a smaller amount,
yielding a natural diminishing-return behavior characteristic of
submodular set functions.

\section{Greedy $(1{-}1/e)$ Guarantee}
\label{app:greedy-guarantee}

In the main paper, we use the notation $\mathrm{EIG}(v\mid S)$ for the
one-step expected information gain of adding a candidate view $v$
after a set of previously selected views $S$, and
$\mathrm{EIG}(S_k)$ for the total information gain accumulated over
$k$ steps of our greedy policy:
\[
\mathrm{EIG}(S_k^{\text{greedy}})
:= \sum_{t=1}^{k}\, \mathrm{EIG}(v_t^{\text{greedy}} \mid S_{t-1}^{\text{greedy}}).
\]

For clarity, we now make explicit how this notation relates to the
adaptive submodularity framework of \cite{golovin2011adaptive}.
Let $F:2^{\mathcal V}\!\to\mathbb{R}$ be the set-level objective
defined by the expected total information gain after observing a set
of views $S_{k}$:
\begin{equation}
F(S_{k})
\;:=\;
\sum_{t=1}^k
  \mathrm{EIG}(v_t \mid S_{t-1}).
\label{eq:set-level-objective}
\end{equation}
In the notation of~\cite{golovin2011adaptive}, $F$ corresponds to the
utility function $f$, and the conditional marginal gain
$\Delta(v\mid\psi)$ under a partial realization $\psi$ is exactly our
one-step $\mathrm{EIG}(v\mid S)$ evaluated after the views in $\psi$.

Lemmas 1 and 2 are stated in terms of $\mathrm{EIG}(v\mid S)$,
but they can equivalently be viewed as establishing that
the set-level objective $F(S)$ is adaptive monotone and
adaptive submodular in the sense of~\cite{golovin2011adaptive}.

Therefore, by Theorem 5 and 16 of~\cite{golovin2011adaptive},
the greedy policy that selects at each step
\(
v_t^{\text{greedy}} = \arg\max_v \mathrm{EIG}(v \mid S_{t-1})
\)
achieves a $(1{-}1/e)$ approximation to the optimal adaptive policy in
terms of the expected total information gain $F(S_k)$:
\begin{equation}
\mathbb{E}\!\left[F(S_k^{\text{greedy}})\right]
\;\ge\;
(1 - 1/e)\,
\max_{\pi}\, \mathbb{E}\!\left[F(S_k^{\pi})\right],
\end{equation}
where $S_k^{\pi}$ denotes the random set of views selected by any
adaptive policy $\pi$.
Since $F(S_k)$ equals $\mathrm{EIG}(S_k)$ by
definition~\eqref{eq:set-level-objective}, this matches the statement
used in the main paper.

\section{Relationship Between Posterior Entropy and Bayes Accuracy}
\label{app:entropy-accuracy}

In our Beta-Bernoulli setting, each Gaussian $i$ has an unknown binary label
$y_i \in \{0,1\}$ with latent probability $p_i = \Pr(y_i=1)$.
Given observations, the posterior distribution of $p_i$ is
\[
    p_i \mid \text{data} \sim \mathrm{Beta}(a_i, b_i),
\]
and the corresponding predictive distribution of the label is Bernoulli with
predictive mean
\[
    q_i := \mathbb{E}[p_i \mid \text{data}] = \frac{a_i}{a_i + b_i}.
\]

The predictive uncertainty of the binary label is measured by the Bernoulli
entropy
\[
    H_{\mathrm{pred}}(q_i)
    = - q_i \log q_i - (1-q_i)\log(1-q_i).
\]
This entropy is maximized at $q_i = 1/2$ and decreases monotonically as $q_i$
moves away from $1/2$.

Under $0$-$1$ loss for each Gaussian, the Bayes-optimal prediction is
\[
    \hat y_i = \begin{cases}
        1, & q_i \ge 1/2,\\
        0, & q_i < 1/2,
    \end{cases}
\]
yielding Bayes accuracy
\begin{align*}
     A_i(q_i)= \max(q_i,\, 1-q_i).
\end{align*}

Because the Bernoulli entropy is symmetric,
$H_{\mathrm{pred}}(q)=H_{\mathrm{pred}}(1-q)$,
it depends only on the smaller of $\{q,1-q\}$.
Hence we can write
\begin{align*}
    H_{\mathrm{pred}}(q)
    &= h\bigl(\min(q,1-q)\bigr), \\
    \quad
    \mathrm{where} \quad h(u)&=-u\log u-(1-u)\log(1-u),
\end{align*}
with $\min(q,1-q)\in[0,1/2]$.
On this interval, the binary entropy satisfies the simple lower bound
\[
    h(u) \;\ge\; 2(\log 2)\,u,
    \qquad 0\le u\le 1/2.
\]
Therefore,
\[
    H_{\mathrm{pred}}(q)
    \;=\;
    h\bigl(\min(q,1-q)\bigr)
    \;\ge\;
    2(\log 2)\,\min(q,1-q),
\]
which implies
\[
    \min(q,1-q)
    \;\le\;
    \frac{H_{\mathrm{pred}}(q)}{2\log 2}.
\]

The per-Gaussian Bayes accuracy is
\[
    A(q)=\max(q,1-q)
        = 1 - \min(q,1-q),
\]
so we obtain the explicit entropy-based lower bound
\begin{align}
    A(q)
    \;\ge\;
    1 - \frac{H_{\mathrm{pred}}(q)}{2\log 2}.
\end{align}

Thus, in our Beta--Bernoulli model, reducing the predictive
entropy $H_{\mathrm{pred}}(q)$ directly tightens a linear lower bound
on the per-Gaussian Bayes-optimal accuracy.

\begin{figure*}[t]
    \centering
    \includegraphics[width=1\textwidth]{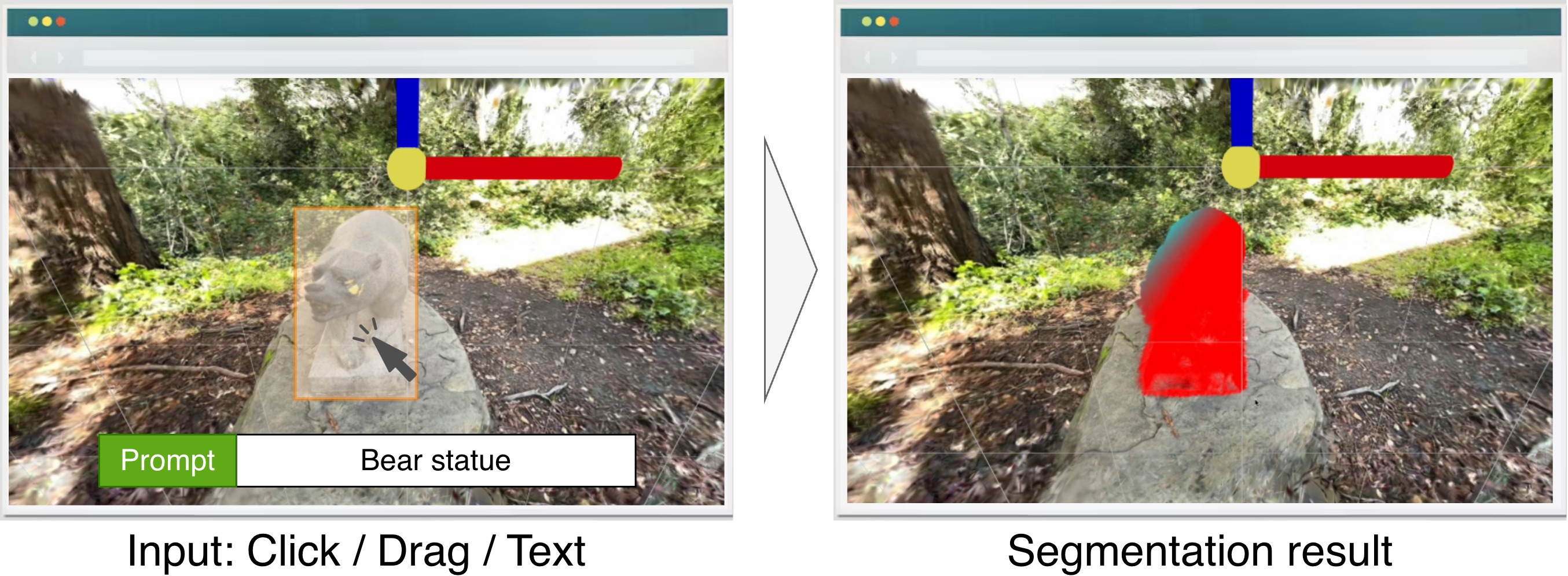}
    \caption{\textbf{Example of assumed application of B$^3$-Seg.}
    A user specifies an object in an interactive 3DGS 
    editor through text, clicking, or dragging.
    Our method then performs fast and stable 3D segmentation from the current
    viewpoint.}
    \label{fig:application}
\end{figure*}

\section{Approximation Gap Between IG and EIG}
\label{app:det-ig-eig}

In the main paper, the information gain (IG) of a candidate view $v$
is defined using the actual 2D mask $M(v)$ returned by the segmentation module:
\begin{align*}
\mathrm{IG}(v)
&=
\sum_i \Bigl[
H(\mathrm{Beta}(a_i,b_i))
\nonumber\\
&\quad -
H\!\left(\mathrm{Beta}(a_i+e_{i,1}(v),\, b_i+e_{i,0}(v))\right)
\Bigr].
\end{align*}
This quantity is deterministic once the view is rendered and its mask has been inferred.

In contrast, our analytic EIG approximates the Beta update
using the posterior mean $m_i=a_i/(a_i+b_i)$:
\begin{align*}
\tilde e_{i,1}(v) &= m_i\tau_i(v),
\quad
\tilde e_{i,0}(v) = (1-m_i)\tau_i(v),
\end{align*}
where $\tau_i(v)=e_{i,1}(v)+e_{i,0}(v)$ is the total responsibility of Gaussian $i$.
Thus,
\begin{align*}
\mathrm{EIG}(v)
&=
\sum_i \Bigl[
H(\mathrm{Beta}(a_i,b_i))
\nonumber\\
&\quad -
H\!\left(\mathrm{Beta}(a_i+\tilde e_{i,1}(v),\, b_i+\tilde e_{i,0}(v))\right)
\Bigr].
\end{align*}

To analyze the difference, define for each Gaussian $i$ the function
\begin{align*}
f_i(w)
=
H\!\left(
\mathrm{Beta}(a_i+w,\; b_i+\tau_i-w)
\right),
\quad w\in[0,\tau_i].
\end{align*}
Then
\begin{align*}
\mathrm{IG}_i(v)
&=
H(\mathrm{Beta}(a_i,b_i))
-
f_i(W_i),
\\
\mathrm{EIG}_i(v)
&=
H(\mathrm{Beta}(a_i,b_i))
-
f_i(\mu_i),
\end{align*}
where $W_i=e_{i,1}(v)$ is the true foreground responsibility and
$\mu_i=m_i\tau_i$ is the posterior-mean prediction.

Hence the approximation error is
\begin{align*}
\Delta_i(v)
&:= \mathrm{IG}_i(v)-\mathrm{EIG}_i(v)= f_i(\mu_i)-f_i(W_i).
\end{align*}

Because $f_i(w)$ is continuously differentiable on the interval $[0,\tau_i]$,
the mean value theorem applies directly.
Therefore, for some $\xi_i \in [\,\min(W_i,\mu_i),\,\max(W_i,\mu_i)\,]$,
\begin{align*}
    f_i(\mu_i)-f_i(W_i)
    &= f_i'(\xi_i)\,\bigl(\mu_i-W_i\bigr).
\end{align*}

Therefore,
\begin{align*}
|\mathrm{IG}_i(v)-\mathrm{EIG}_i(v)| 
&\le
\left(
\sup_{u\in[0,\tau_i]} |f_i'(u)|
\right)
|\mu_i-W_i|.
\end{align*}

The term $|\mu_i-W_i|=\tau_i\,|m_i-W_i/\tau_i|$
reflects the mismatch between the posterior belief $m_i$ and the true proportion
$W_i/\tau_i$ inferred by the mask.
As the Beta posterior becomes aligned with the observed masks, this mismatch
shrinks, and so does the approximation error.

Thus, once a few updates have been incorporated, the analytic EIG becomes a
highly accurate surrogate of the true IG while requiring no mask inference for
candidate views, explaining the strong correlation observed in Fig. \ref{fig:scatter_pred_vs_actual}.

\begin{figure*}[t]
    \centering
    \includegraphics[width=0.9\textwidth]{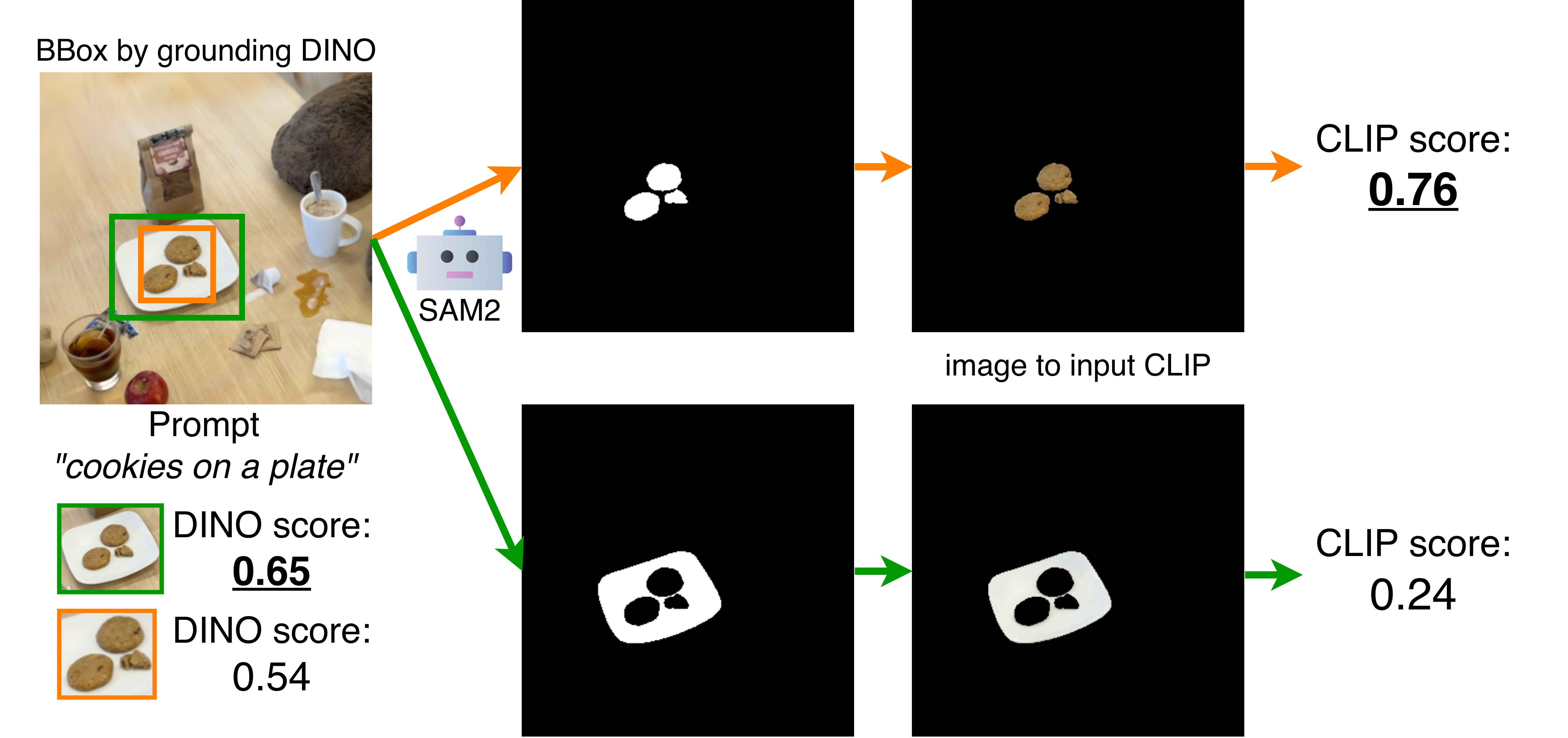}
    \caption{Effect of CLIP re-ranking in the LERF-Mask Teatime scene.
Although GroundingDINO assigns a higher score to the wrong bounding box (green),
CLIP correctly assigns a higher similarity score to the region corresponding to the
true object described by the prompt ``cookies on a plate’’ (orange).}
    \label{fig:effect_of_clip}
\end{figure*}

\section{Assumed Application}
\label{app:application}

A primary motivation behind our method is the design of 
interactive 3DGS editing systems as web applications.
In such an application, a user inspects a scene from an arbitrary
camera viewpoint and selects a target object via text prompts
(e.g., ``bear statue''), clicks or drag gestures 
(Fig.~\ref{fig:application}).
The system must immediately infer the corresponding 3D region, 
often within a fraction of a second, before further editing or manipulation
(color changes, deletion, mesh extraction, etc.) can proceed.

This setting naturally provides a canonical initial view:  
the viewpoint from which the user initiates the interaction.
Because the user always begins by examining the scene visually,
an initial camera view for the initial object mask is available by design.

Thus, the assumption of an initial canonical view in our method is not an
artificial restriction, but a realistic property of interactive 3DGS
workflows.  
In fact, B$^3$-Seg is tailored for this usage scenario: its closed-form EIG,
adaptively diminishing uncertainty, and $\sim$12s end-to-end inference make it
particularly suitable for real-time editing environments.

\section{Effectiveness of CLIP re-ranking}
Text-driven region proposal by GroundingDINO is often coarse and sometimes
selects a bounding box that does not match the user query with high confidence.
To verify that CLIP re-ranking can correct such failure cases,
we performed an ablation on the LERF-Mask Teatime scene using the prompt
\textit{``cookies on a plate''}.
Figure~\ref{fig:effect_of_clip} shows a representative example.

GroundingDINO produces two candidate bounding boxes for this prompt.
Despite the lower DINO score, the orange box corresponds to the correct region,
while the green box (incorrect) receives a higher DINO confidence
(0.65 vs.\ 0.54).
We apply SAM2 to each bounding box to obtain a candidate segmentation mask,
crop the corresponding region, and compute the CLIP image–text similarity.

CLIP successfully assigns substantially higher similarity to the correct region, reversing the erroneous DINO ordering.
This confirms that CLIP acts as a strong fine-grained verifier for
text–region alignment, and that the two-stage
DINO → SAM2 → CLIP re-ranking pipeline significantly improves
the reliability of text-guided segmentation.

\section{Future Directions}
\subsection{Multi-class Segmentation}
In this work, we focused on binary foreground--background decisions with a Beta--Bernoulli model for each Gaussian. A natural extension is to support multi-class segmentation by replacing the Beta--Bernoulli pair with a Dirichlet--Categorical model.

Concretely, let each Gaussian label take values in a finite set of classes $\{1,\dots,K\}$. We introduce per-Gaussian class probabilities
\[
    \boldsymbol{\pi}_i = (\pi_{i,1},\dots,\pi_{i,K}), \qquad
    \sum_{c=1}^K \pi_{i,c} = 1,
\]
and place a Dirichlet prior/posterior on $\boldsymbol{\pi}_i$:
\[
    y_i \mid \boldsymbol{\pi}_i \sim \mathrm{Categorical}(\boldsymbol{\pi}_i),
    \qquad
    \boldsymbol{\pi}_i \sim \mathrm{Dirichlet}(\boldsymbol{a}_i),
\]
where $\boldsymbol{a}_i = (a_{i,1},\dots,a_{i,K})$ denotes the class-wise pseudo-counts.

Given a rendered view $v$ and a multi-class mask, we aggregate class-specific evidence for each Gaussian in analogy to Eq.~\eqref{eq:agg-evidence}. Let $e_{i,c}(v)$ denote the responsibility of Gaussian $g_i$ assigned to class $c$ in view $v$ (e.g., via per-pixel class masks or soft class scores):
\[
    e_{i,c}(v)
    \;=\;
    \sum_{(j,k)\in I(v)} \alpha_i T_i \,\mathbb{I}[M_{j,k}(v)=c],
    \quad c\in\{1,\dots,K\}.
\]
The binary Beta update in Eq.~\eqref{eq:beta-update} is then replaced by the Dirichlet update
\[
    \mathrm{Dirichlet}(\boldsymbol{a}_i)
    \;\leftarrow\;
    \mathrm{Dirichlet}\bigl(a_{i,1}{+}e_{i,1}(v),\dots,a_{i,K}{+}e_{i,K}(v)\bigr),
\]
and after multiple views we obtain
\[
    \boldsymbol{\pi}_i
    \sim
    \mathrm{Dirichlet}\Bigl(
        a_{\mathrm{init}}{+}\textstyle\sum_v e_{i,1}(v),\dots,
        a_{\mathrm{init}}{+}\textstyle\sum_v e_{i,K}(v)
    \Bigr).
\]
The posterior mean $\mathbb{E}[\pi_{i,c}] = \alpha_{i,c}/\sum_{c'}\alpha_{i,c'}$ then replaces the Beta mean used for binary decisions, enabling multi-class 3D labeling.

The information-theoretic view of Sec.~\ref{sec:eig} also extends naturally. The Beta entropy $H(\mathrm{Beta}(a_i,b_i))$ in Eq.~\eqref{eq:information-gain} and Eq.~\eqref{eq:eig} is replaced by the Dirichlet entropy $H(\mathrm{Dirichlet}(\boldsymbol{a}_i))$, and the per-view information gain becomes
\[
    \mathrm{IG}(v)
    = \sum_i \Bigl[
        H\bigl(\mathrm{Dirichlet}(\boldsymbol{a}_i)\bigr)
        - H\bigl(\mathrm{Dirichlet}(\boldsymbol{a}_i + \mathbf{e}_i(v))\bigr)
      \Bigr],
\]
where $\mathbf{e}_i(v) = (e_{i,1}(v),\dots,e_{i,K}(v))$.
The analytic EIG in Eq.~\eqref{eq:eig} can analogously be defined by replacing the Beta mean $m_i$ and entropy with their Dirichlet counterparts, using the posterior mean
$m_{i,c} = \alpha_{i,c}/\sum_{c'}\alpha_{i,c'}$ to construct approximate evidence $\tilde e_{i,c}(v)$ for each class as follows.

\[
\tilde e_{i,c}(v) = m_{i,c}\tau_i = \sum_{(j,k) \in I(v)} \alpha_i T_i \frac{a_{i,c}}{\sum_{c'}a_{i,c'}}
\]
Thus EIG is
\[
    \mathrm{EIG}(v)
    = \sum_i \Bigl[
        H\bigl(\mathrm{Dirichlet}(\boldsymbol{a}_i)\bigr)
        - H\bigl(\mathrm{Dirichlet}(\boldsymbol{a}_i + \tilde{\mathbf{e}}_i(v))\bigr)
      \Bigr].
\]

In this Dirichlet-Categorical setting, the concentration parameter generalizes to
$\kappa_i = \sum_{c=1}^K \alpha_{i,c}$, and the same intuition about monotonicity and diminishing returns carries over: increasing $\kappa_i$ while keeping the class proportions fixed makes the posterior more concentrated and reduces the entropy. We therefore expect the adaptive monotonicity and adaptive submodularity arguments of Sec.~\ref{sec:theory} to extend to the multi-class regime, yielding a principled path toward fully multi-object, open-vocabulary 3DGS segmentation within the same EIG-driven framework.

\subsection{Entropy-Based Early Stopping}
The results in Appendix~\ref{app:entropy-accuracy} imply that the predictive
entropy of the Beta--Bernoulli posterior provides a quantitative lower bound
on the achievable Bayes accuracy.  
For a Gaussian with posterior predictive probability $q_i$, the Bernoulli
entropy $H_{\mathrm{pred}}(q_i)$ satisfies
\[
    A_i(q_i) \;\ge\;
    1 - \frac{H_{\mathrm{pred}}(q_i)}{2\log 2},
\]
where $A_i(q_i)$ denotes the Bayes-optimal per-Gaussian accuracy.  
Thus, decreasing the entropy directly tightens a guaranteed lower bound on the
final segmentation accuracy.

This observation enables a principled \emph{entropy-based early stopping}
criterion for our sequential inference pipeline.
Let
\[
    \bar H_t := \frac{1}{N}\sum_{i=1}^{N} H_{\mathrm{pred}}(q_i^{(t)})
\]
denote the average predictive entropy at iteration $t$.
From the inequality above, the average Bayes accuracy after iteration $t$ is
lower bounded as
\[
    \bar A_t \;\ge\; 1 - \frac{\bar H_t}{2\log 2}.
\]
Therefore, instead of fixing the number of iterations to a constant $T$, one
can run the view-selection loop until $\bar H_t$ falls below a user-specified
threshold $\bar H_{\mathrm{target}}$, corresponding to a desired guaranteed
accuracy level
\[
    \bar A_{\mathrm{target}}
    \;=\;
    1 - \frac{\bar H_{\mathrm{target}}}{2\log 2}.
\]

Such an adaptive stopping rule ensures that inference terminates once the
posterior has become sufficiently concentrated, avoiding unnecessary queries
while maintaining provable segmentation quality.
Integrating this early-stopping mechanism with analytic EIG is a promising
future direction for further improving the responsiveness and efficiency of
B$^3$-Seg.

\section{Additional Qualitative Results}

\begin{figure*}[t]
    \centering
    \includegraphics[width=\linewidth]{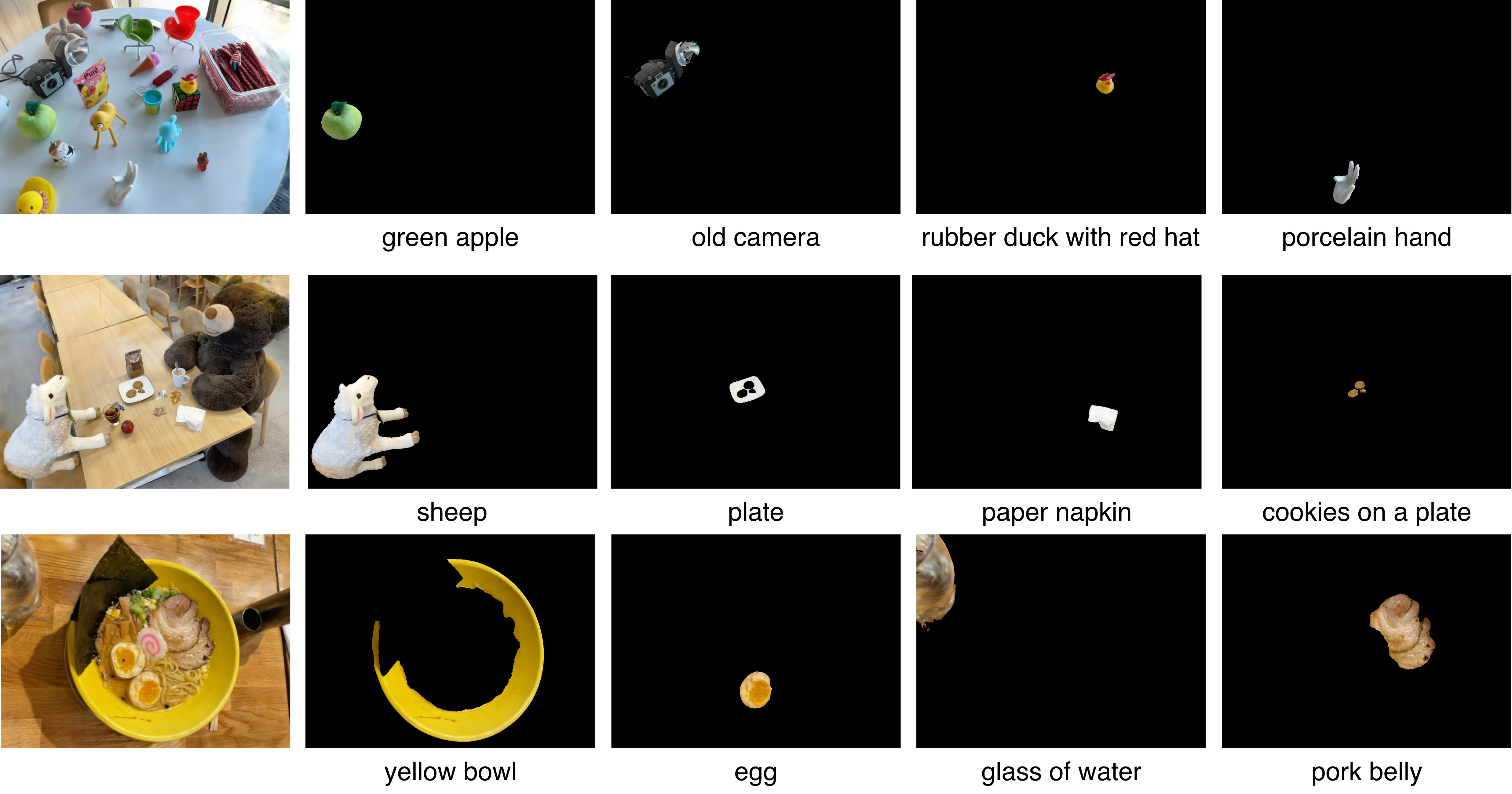} 
    \caption{Qualitative results of B$^{3}$-Seg on LERF-Mask}
    \label{fig:detail_lerf}
\end{figure*}

\begin{figure*}[t]
    \centering
    \includegraphics[width=\linewidth]{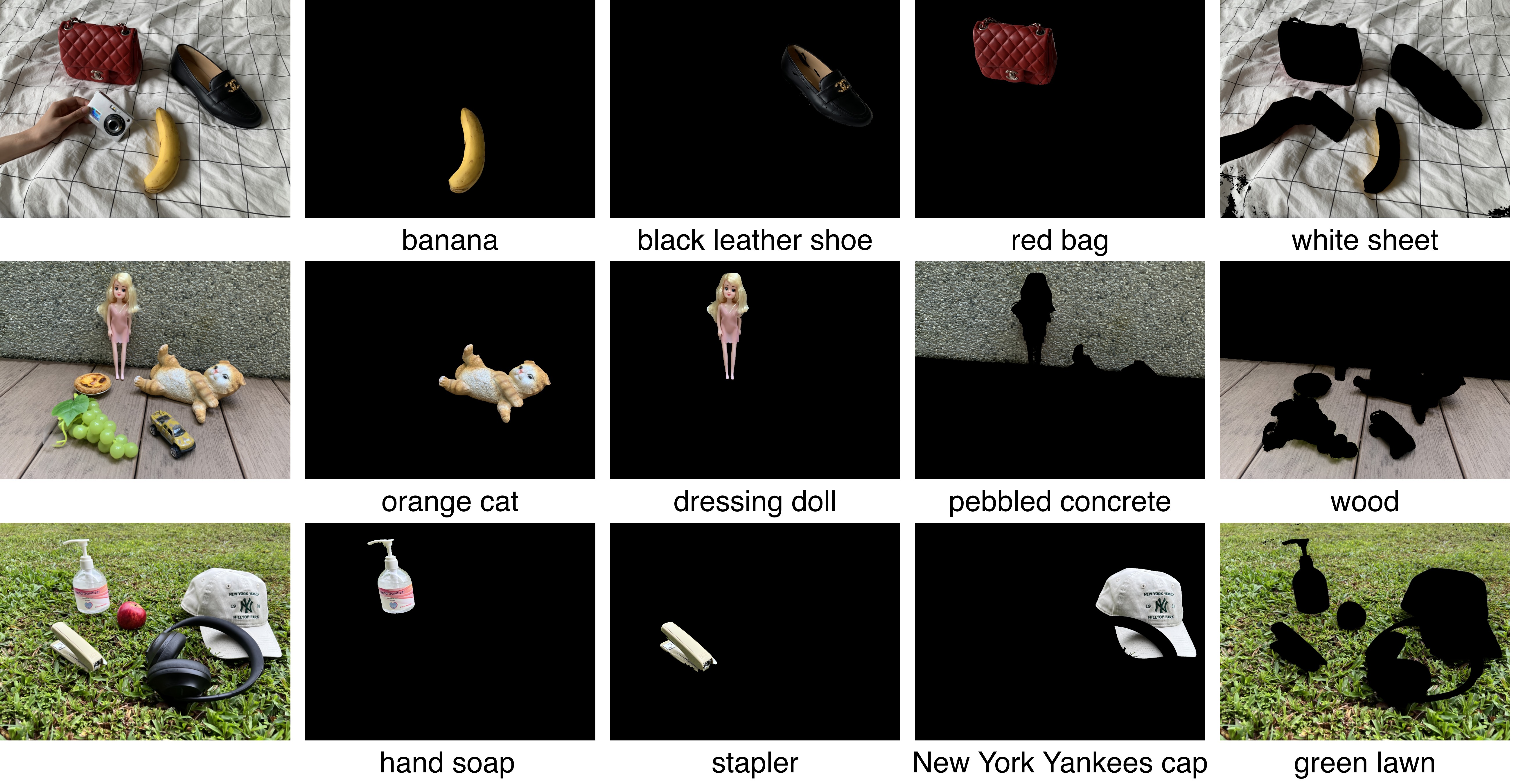} 
    \caption{Qualitative results of B$^{3}$-Seg on 3D-OVS}
    \label{fig:detail_3dovs}
\end{figure*}

\begin{figure*}[t]
    \centering
    \includegraphics[width=0.95\linewidth]{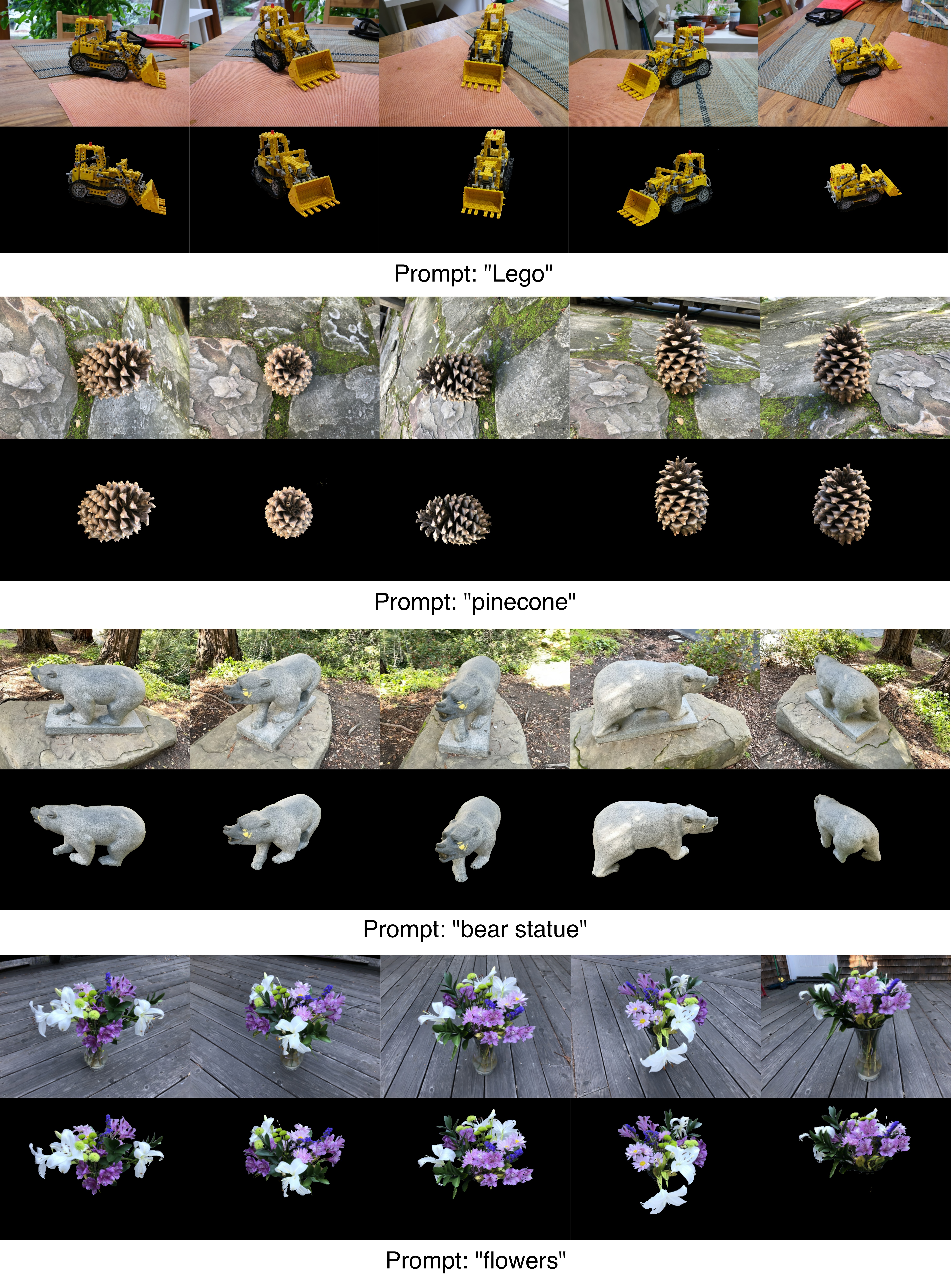} 
    \caption{Qualitative results of other 3D scenes}
    \label{fig:other360}
\end{figure*}

To further illustrate the generality and robustness of B$^{3}$-Seg, 
Figures~\ref{fig:detail_lerf}, \ref{fig:detail_3dovs}, and \ref{fig:other360} present 
additional qualitative results on each 3D dataset. 
Across a wide variety of object types—ranging from small items 
(e.g., fruit, plates, napkins) to deformable or texture-rich objects 
(e.g., stuffed animals, clothing, wooden materials), our method 
generally produces clean and spatially coherent 3D masks.

Notably, B$^{3}$-Seg successfully handles objects with thin structures, 
strong self-occlusion, low contrast against the background, and cases 
where multiple distractors appear in close proximity. These results 
highlight that (i) analytic EIG selects informative viewpoints even in 
cluttered scenes, and (ii) the sequential Beta–Bernoulli updates 
accumulate stable evidence despite imperfect 2D masks. 
\newpage

\newpage

\end{document}